%% file: main.tex
\pdfoutput=1
\documentclass[letterpaper]{article} % DO NOT CHANGE THIS
\usepackage{aaai24}  % DO NOT CHANGE THIS
\usepackage{times}  % DO NOT CHANGE THIS
\usepackage{helvet}  % DO NOT CHANGE THIS
\usepackage{courier}  % DO NOT CHANGE THIS
\usepackage[hyphens]{url}  % DO NOT CHANGE THIS
\usepackage{graphicx} % DO NOT CHANGE THIS
\usepackage{natbib}  % DO NOT CHANGE THIS AND DO NOT ADD ANY OPTIONS TO IT
\usepackage{caption} % DO NOT CHANGE THIS AND DO NOT ADD ANY OPTIONS TO IT
\usepackage{algorithm}
\usepackage{algorithmic}
\usepackage{amsmath}
\usepackage{booktabs} 
\usepackage{multirow}
\usepackage{wasysym}
\usepackage{amssymb}
\usepackage{appendix}
\usepackage{newfloat}
\usepackage{listings}
\DeclareCaptionStyle{ruled}{labelfont=normalfont,labelsep=colon,strut=off} % DO NOT CHANGE THIS
\floatstyle{ruled}
\newfloat{listing}{tb}{lst}{}
\floatname{listing}{Listing}
\title{PPEA-Depth: Progressive Parameter-Efficient Adaptation for \\Self-Supervised Monocular Depth Estimation}
\author {
    Yue-Jiang Dong\textsuperscript{\rm 1},
    Yuan-Chen Guo\textsuperscript{\rm 1}, 
    Ying-Tian Liu\textsuperscript{\rm 1},
    Fang-Lue Zhang\textsuperscript{\rm 2}, 
    Song-Hai Zhang\textsuperscript{\rm 1}
    \thanks{Project homepage: https://yuejiangdong.github.io/PPEADepth/.}
    \thanks{Corresponding author.}
}
\affiliations {
    \textsuperscript{\rm 1}BNRist, Department of Computer Science and Technology, Tsinghua University, China \\
    \textsuperscript{\rm 2}Victoria University of Wellington, New Zealand\\
    \{dongyj21@mails., guoyc19@mails., liuyingt23@mails., shz@\}tsinghua.edu.cn, fanglue.zhang@vuw.ac.nz
}

\begin{document}

\maketitle

\begin{abstract}
Self-supervised monocular depth estimation is of significant importance with applications spanning across autonomous driving and robotics. However, the reliance on self-supervision introduces a strong static-scene assumption, thereby posing challenges in achieving optimal performance in dynamic scenes, which are prevalent in most real-world situations.
To address these issues, we propose PPEA-Depth, a Progressive Parameter-Efficient Adaptation approach to transfer a pre-trained image model for self-supervised depth estimation. The training comprises two sequential stages: an initial phase trained on a dataset primarily composed of static scenes, succeeded by an expansion to more intricate datasets involving dynamic scenes. To facilitate this process, we design compact encoder and decoder adapters to enable parameter-efficient tuning, allowing the network to adapt effectively. They not only uphold generalized patterns from pre-trained image models but also retain knowledge gained from the preceding phase into the subsequent one. Extensive experiments demonstrate that PPEA-Depth achieves state-of-the-art performance on KITTI, CityScapes and DDAD datasets.
\end{abstract}

\input{data/1introduction}
\input{data/2relatedwork}
\input{data/3approach}

\input{data/4experiments}

\input{data/5conclusion}

\section*{Acknowledgements}
This work was supported by the National Key Research and Development Program of China (No. 2023YFF0905104),
the Natural Science Foundation of China (No. 62132012), Beijing Municipal Science and Technology Project (No. Z221100007722001) and Tsinghua-Tencent Joint Laboratory for Internet Innovation Technology. Fang-Lue Zhang was supported by the Marsden Fund Council managed by the Royal Society of New Zealand (No. MFP-20-VUW-180).

\newpage

\appendix
\section*{Appendix}

\input{data/a0exp}
\input{data/a1ablations}

\input{data/a2latency}
\input{data/a3details}
\input{data/a4qualitative}

\end{document}

%% file: data/1introduction.tex
\section{Introduction}
\begin{figure}[t!]
    \centering
    \includegraphics[width=\linewidth]{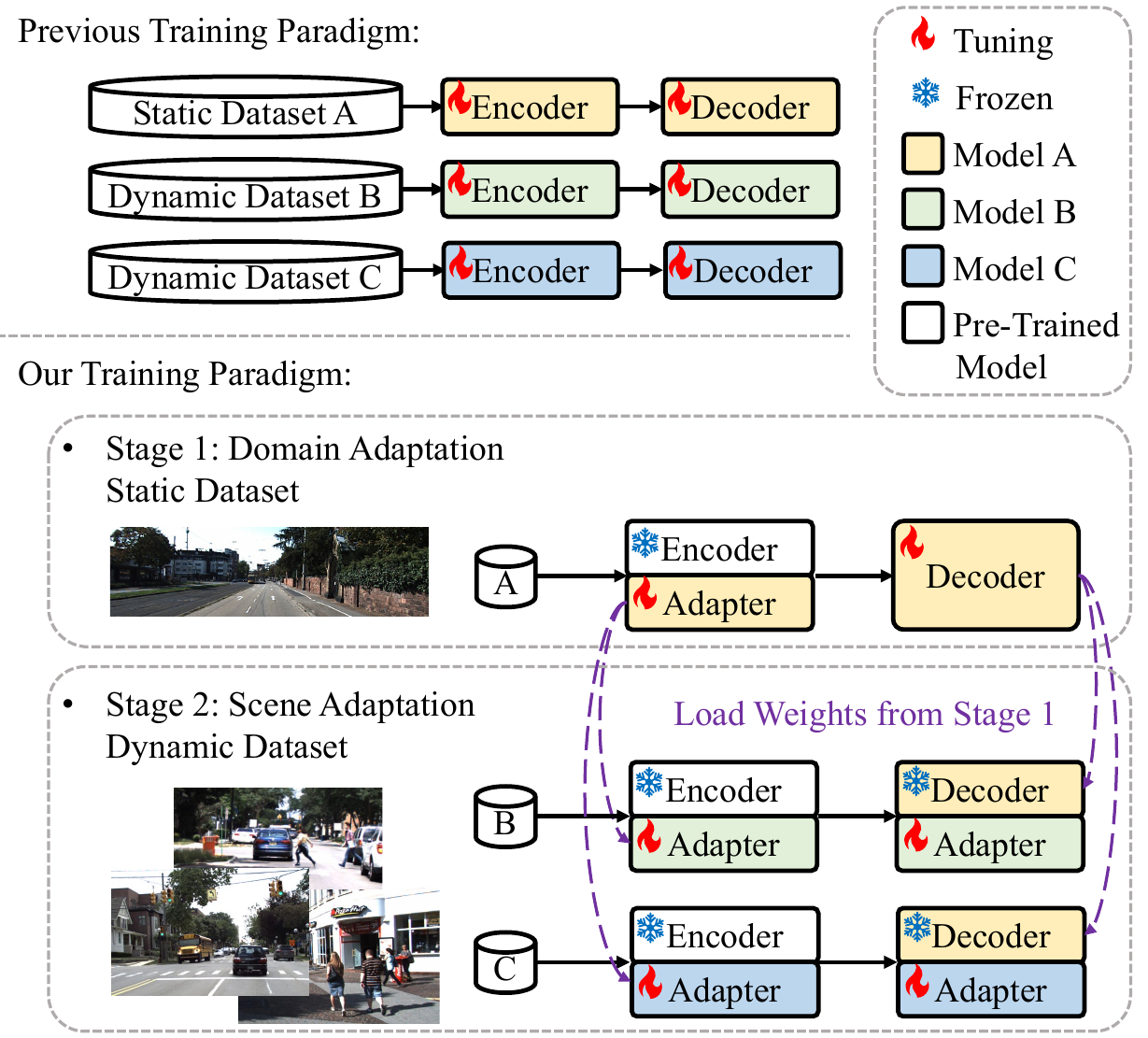}
    \caption{\textbf{Previous Paradigm v.s. Our Paradigm}. The conventional training approach employs a consistent process for both static and dynamic datasets: it includes using a pre-trained image model as an encoder and fine-tuning all U-Net parameters for each dataset. In contrast, our novel two-stage training paradigm integrates adapters to progressively tailor the pre-trained image models for depth perception initially on simple datasets (static scenes primarily) and then extends to intricate datasets (with dynamic scenes).}
    \label{fig:poster}
\end{figure}
In the realm of computer vision, accurate depth perception of a scene is a fundamental aspect that underpins a wide array of applications, from autonomous vehicles navigating complex environments to immersive virtual reality experiences. Depth estimation has witnessed remarkable advancements in recent years due to the proliferation of deep learning techniques. Especially, self-supervised methods \cite{Zhou_Brown_Snavely_Lowe_2017, monodepth2, watson2021temporal,guizilini2022multi,bangunharcana2023dualrefine} that leverage the inherently contained rich and diverse sources of depth-related information in monocular videos open the door to scalability and real-world adaptability for depth estimation methods. 

Existing self-supervised monocular depth estimation methods are commonly based on fine-tuning pre-trained image models learned from large image datasets, such as ImageNet \cite{deng2009imagenet}, on depth datasets, such as KITTI \cite{geiger2012we}, improving depth estimation accuracy compared to training from scratch \cite{monodepth2}. Some recent approaches introduce cost volume construction within the network to incorporate multi-frame inputs during inference \cite{watson2021temporal,guizilini2022multi,bangunharcana2023dualrefine}. However, the above methods are all rooted in the assumption of a static scene, where solely the camera moves. This strong assumption subsequently creates challenges for these methods to reach optimality in dynamic scenes, limiting the efficient utilization of actual, unlabeled data for self-supervised training. Some other methods propose sophisticated algorithms that incorporate supplementary components like semantic segmentation or motion prediction networks to model object motion \cite{klingner2020self,lee2021learning,feng2022disentangling,hui2022rm}. However, these methods often yield less satisfactory outcomes in static scenes compared to dedicated static methods \cite{guizilini2022multi,bangunharcana2023dualrefine}.

In this paper, we aim to provide a new learning paradigm for self-supervised depth estimation to improve the performance and generalizability of the model, specifically by equipping it with improved capabilities to handle dynamic scenes. During our preliminary experiments of directly fine-tuning pre-trained models using such videos, we observed that excessive training with a relatively small dataset can disrupt the generalized patterns learned during pre-training, potentially resulting in catastrophic forgetting. Inspired by the recent success of parameter-efficient fine-tuning (PEFT) in natural language processing (NLP) community \cite{houlsby2019parameter,pfeiffer2020adapterfusion,hu2021lora} and image and video classification \cite{jia2022visual,chen2022vision, lin2022frozen,chen2022adaptformer,yang2023aim}, we extend it to self-supervised depth estimation, a loose-contrained regression task that remains unexplored in this field. We not only investigate adapters on encoders~\cite{houlsby2019parameter} to improve the adaptation from pre-trained models to the depth perception task, but also propose a novel decoder adapter to boost network robustness towards dynamic scenes. 

We propose an innovative two-stage self-supervised depth estimation approach, PPEA-Depth, based on parameter-efficient adaptation. We devise lightweight encoder adapters and decoder adapters within our framework.  
Our method is guided by insights drawn from human learning mechanisms, which typically progress from simple to complex tasks.
Our approach first trains on datasets primarily featuring static scenes, which adhere to the static scene assumptions. The pre-trained encoder is frozen to retain general patterns gained from large image datasets and encoder adapters are tuned to adapt the network to learn depth priors.
Subsequently, we train the network on more intricate and dynamic scenes. To retain the knowledge learned from static scenes in the preceding stage, we load weights from the previous stage, freeze both the encoder and decoder, and just train extra encoder and decoder adapters to summarize network updates for adaptation to new scenes. 

The second scene adaptation stage extends beyond a single dataset. The domain-adapted model can be flexibly adapted to various new scenes solely by tuning the adapters. We only need to tune and store a small number of scene-specific parameters to generalize across different datasets. With these innovations, PPEA-Depth achieves state-of-the-art performance on KITTI, CityScapes\cite{cordts2016cityscapes} and DDAD\cite{guizilini20203d} datasets. 

Our main contributions can be summarized as:
\begin{itemize}
    \item We propose a new paradigm to transfer upstream pre-trained models to self-supervised monocular depth estimation in a progressive manner from static scenes to more challenging dynamic scenes.
    \item We design encoder adapters to take advantage of pre-trained image models. Reducing tunable parameters by up to 90\%, tuning encoder adapters demonstrates less depth estimation errors than full fine-tuning.
    \item We design a decoder adapter to enhance the adaptability of the decoder to more challenging datasets. Remarkably, merely tuning encoder adapters and the decoder adapter yields a 6\% improvement in absolute relative errors compared to fine-tuning all U-Net parameters on the full training set when utilizing only 3\% of the training data. 
\end{itemize}

%% file: data/2relatedwork.tex
\section{Related Work}

\subsection{Self-Supervised Monocular Depth Estimation} 
Self-supervised monocular depth estimation predicts depth and camera ego-motion from an outdoor monocular video, and is supervised by image reprojection loss \cite{Zhou_Brown_Snavely_Lowe_2017}. On the basis of such methodology, previous works make progress in designing loss functions for better convergence to optimum \cite{monodepth2,shu2020feature}, designing more complicated encoder structure with cost volume \cite{watson2021temporal,bangunharcana2023dualrefine} and attention scheme \cite{guizilini2022multi}, 
and leveraging cross-domain information of optical flow \cite{yin2018geonet,chen2019self,ranjan2019competitive}  or scene semantics \cite{casser2019unsupervised,klingner2020self,jung2021fine,lee2021learning,lee2021attentive,feng2022disentangling} to handle dynamic objects, etc. Training self-supervised depth estimation from a pre-trained model yields superior performance compared to training from scratch \cite{monodepth2}, implying the generalized patterns learned in pre-training benefit this task.

\subsection{Parameter-Efficient Fine-Tuning}
% PEFT is proposed in NLP tasks
A variety of parameter-efficient fine-tuning (PEFT) methods have been proposed in recent NLP works \cite{houlsby2019parameter,pfeiffer2020adapterfusion,karimi2021compacter, zaken2021bitfit,zhu2021counter, li2021prefix, hu2021lora,he2021towards}. Different from the traditional training patterns which fine-tune large pre-trained models on different downstream tasks, PEFT freezes parameters in pre-trained models and only fine-tunes a small number of extra parameters to obtain strong performance with less tuned parameters.

% PEFT in CV tasks
In the computer vision field, PEFT has been mainly studied and applied in classification tasks including image classification \cite{jia2022visual,bahng2022visual,chen2022adaptformer,jie2023fact} and video action recognition \cite{lin2022frozen,chen2022adaptformer,yang2023aim}. 
% vision transformer dense prediction
A recent work \cite{chen2022vision} investigates PEFT for the application of Vision Transformer \cite{dosovitskiy2020image} in dense prediction tasks including semantic segmentation and object detection.  Different from previous work, we study PEFT in self-supervised monocular depth estimation, a challenging dense regression task. The PEFT algorithm is not only designed and applied for parameters of the encoder backbone but also for the dense prediction decoder in our method. To the best of our knowledge, we are the first to study PEFT in the depth estimation area.

\begin{figure*}[t!]
    \centering
    \includegraphics[width=\textwidth]{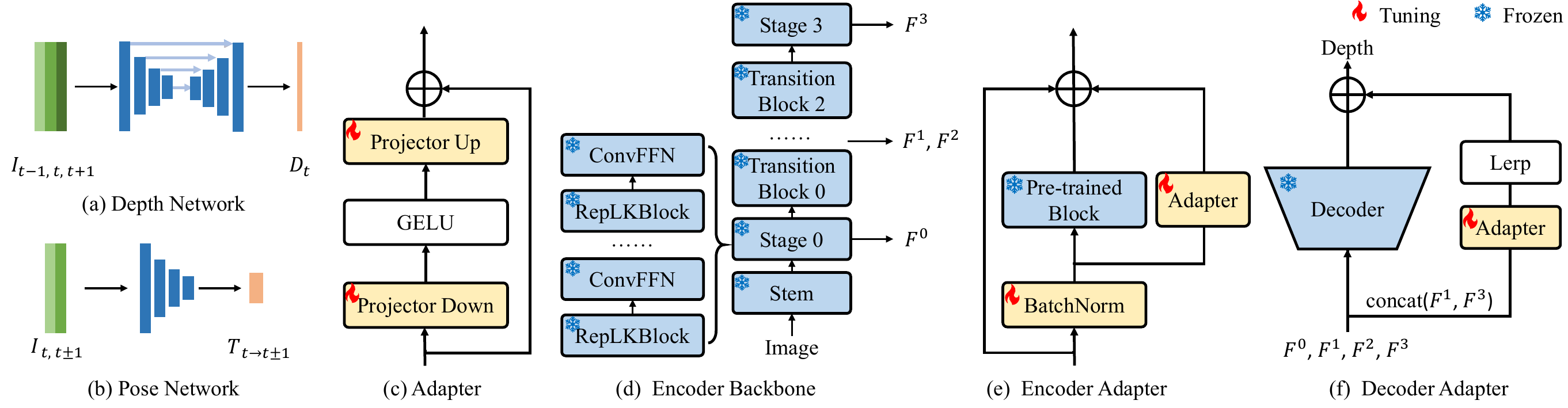}
    \caption{(a) Depth network is a U-Net structure predicting depth taking three consecutive frames. (b) Pose network regresses the camera relative pose given two images. (c) Adapter is a bottleneck structure with a skip connection. (d) Structure of RepLKNet \cite{replknet} backbone. (e) Our encoder adapter design. We attach encoder adapters to pre-trained RepLKBlock and ConvFFN. (f) Our decoder adapter design. Lerp represents linear interpolation.}
    \label{fig:overview}
\end{figure*}

%% file: data/3approach.tex
\section{Method}

\subsection{Overview}
Our network comprises a depth network (Fig. \ref{fig:overview}(a)) and a pose network (Fig. \ref{fig:overview}(b)). The depth network employs a U-Net structure, encompassing an encoder to extract image features and a decoder to predict dense depth maps. Meanwhile, the pose network predicts the camera transformation between two frames. It has a feature extractor followed by a prediction head, which outputs a six-dimensional vector – three for rotation angles and the other three for translation. 

Our network takes three consecutive frames from a monocular video as input. The middle frame is reconstructed with its adjacent frames, and the difference between the reconstructed and original images serves as the supervision signal. This reconstruction relies on cross-frame pixel correspondences in the structure-from-motion theory \cite{Zhou_Brown_Snavely_Lowe_2017}. Given camera intrinsics $K$ and camera relative pose $T$ between two frames $a, b$ from a video, the pixel correspondence between them can be computed by:
\begin{equation}\label{formula:warp}
    p_{a}\sim KT_{b\to a}D_{b}(p_{b})K^{-1}p_{b},
\end{equation}
where $p_{a}$ and $p_{b}$ are corresponding pixels in the two frames, and $D_b$ is the depth of $p_b$. Assuming a static scene, these correspondences reconstruct frame $b$ from pixels in frame $a$.

We adopt the identical architecture and training strategy for the pose network as ManyDepth \cite{watson2021temporal}, with our primary focus centered on the depth network. In the conventional training approach, encoders are initialized with transferred pre-trained weights, while decoders and prediction heads are trained from scratch. In our method, we introduce the PEFT scheme by adding adapters \cite{houlsby2019parameter} to the depth network's encoder and decoder to encapsulate network adaptations across distinct domains. 

\subsection{Adapter}
Adapters are compact architectures designed to tailor a pre-trained module for a specific downstream task \cite{houlsby2019parameter}. While parameters of the pre-trained model remain static, only adapter parameters are modified during training. Illustrated in Fig. \ref{fig:overview}(c), adapters follow a bottleneck structure, encompassing two linear projection layers, an activation layer, and a skip connection. The initial projection layer reduces the input feature dimension, and the subsequent one restores it to the original input dimension after the activation layer.
Based on such architecture, we respectively design adapters for the depth network encoder and decoder.

\subsection{Encoder Adapter}
\subsubsection{Backbone} 
We opt for RepLKNet \cite{replknet}, a CNN architecture featuring a notable kernel size of $31\times 31$, as the encoder backbone. This selection is attributed to its adaptability concerning input image resolution, comparable accuracy to Swin Transformer \cite{liu2021Swin}, and enhanced inference speed when applied to downstream tasks. 

As illustrated in Fig. \ref{fig:overview}(d), RepLKNet generates feature maps of four different scales: $F^{1},F^{2},F^{3},F^{4}$ at four stages. Within each stage, a RepLKBlock and a ConvFFN are interleaved in their arrangement. Please refer to the supplementary materials for more details. To leverage the pre-established generalized patterns of the RepLKNet and fine-tune them for the depth regression task, we integrate adapters to the RepLKBlocks and ConvFFNs. 

\subsubsection{Architecture} 
As depicted in Fig. \ref{fig:overview}(e), the input feature is initially processed through a batch normalization layer and then fed into the pre-trained block and adapter using two parallel streams. Within each adapter, we adhere to the conventional bottleneck structure, with a notable distinction being the replacement of the first linear projection module with a convolutional layer for the adapters of RepLKBlock. This convolutional layer employs a kernel size of 3, a stride of 1, and padding of 1. These settings maintain the spatial dimensions of the input feature maps unchanged after convolution. Substituting the linear projection with a $3\times 3$ convolution offers adapters a larger receptive field, proving advantageous for per-pixel regression tasks such as depth estimation. Meanwhile, the ConvFFN's adapter continues to adopt linear projection in order to minimize the parameter count. Given the input $x$, the output $x'$ of the module after incorporating the adapted block $\mathcal{A}$ can be written as:
\begin{equation}x' = x + \mathcal{M}(\mathcal{N}(x))  + \mathcal{A}(\mathcal{N}(x)),
\end{equation}
where $\mathcal{M}$ is the pre-trained block, $\mathcal{A}$ is its adapter, and $\mathcal{N}$ represents batch normalization.

\subsection{Decoder Adapter}
The architecture of the decoder adapter is illustrated in Fig. \ref{fig:overview}(f). To strike a balance between the additional parameter count and the sufficiency of adapter input, we perform an interpolation of $F^{3}$ to match the spatial dimensions of $F^{0}$, and subsequently use their concatenation as the input to the adapter. The inner structure of the decoder adapter adopts a \textit{Projector-GELU-Projector} configuration, where both projectors employ linear projection. Given that $F^{0}$ has dimensions of $(H/4, W/4)$ spatially, the output of the decoder adapter requires restoration to the original size of the input images, i.e., $(H, W)$. This is achieved through linear interpolation. The computation for incorporating the decoder adapter can be written as:
\begin{equation}
    x' = \mathcal{D}(F^{0}, F^{1}, F^{2}, F^{3}) + \mathcal{A}(F^{0}, F^{3}),
\end{equation}
where $\mathcal{D}$ is the decoder and $\mathcal{A}$ is its adapter.

\subsection{Progressive Adaptation}
As illustrated in Fig. \ref{fig:poster}, our progressive adaptation involves two stages. Stage 1 is trained on a dataset that primarily follows the static-scene assumption. In Stage 1, a pre-trained model, capable of efficiently extracting color features, is tailored from an image classification task to depth regression. We freeze all encoder parameters and only train encoder adapters and U-Net decoder. The frozen encoder retains the generalized patterns acquired during the model pre-training.

Stage 2 is conducted on datasets that predominantly feature dynamic scenes, which are more challenging for training because dynamic objects violate Eqn. (\ref{formula:warp}) and mislead the self-supervision signal. 
Our method capitalizes on the depth priors obtained from static scenes in Stage 1 and applies them to dynamic scenes. 
In Stage 2, we load the weights of the U-Net encoder, the encoder adapters, and the U-Net decoder from Stage 1, and freeze both the encoder and decoder, with only adapter parameters being updated. This paradigm preserves the depth perception ability obtained from Stage 1, as most network parameters are frozen and are unaffected by the erroneous loss caused by object motion. Meanwhile, the lightweight adapters make minor adjustments based on this robust depth prior, fitting data distribution in new scenes.

%% file: data/4experiments.tex
\section{Experiments}
In this section, we (1) demonstrate the effectiveness of the two stages separately, (2) showcase that PPEA-Depth yields state-of-the-art results on standard benchmarks, and (3) assess the generalizability of the proposed methods. Supplementary materials contain additional results and ablation studies to validate our approach.
\begin{table*}[h]
    \centering
    % \small
    \scalebox{0.9}{
    \begin{tabular}{ccccccccccccc}
        \toprule
        {\multirow{2}{*}{Pre-Trained}} &\multicolumn{3}{l}{\multirow{2}{*}{Tuning Strategy}} {}& \multicolumn{2}{c}{{Tuning Params (M)}} & \multicolumn{4}{c}{\emph{Errors}$\downarrow$} & \multicolumn {3}{c}{\emph{Accuracy}$\uparrow$}\\
        \cmidrule(r){5-6} \cmidrule(r){7-10} \cmidrule(r){11-13}
        {Backbone} &  & & &{Encoder} & Decoder & AbsRel & SqRel & RMSE & RMSE$_\mathrm{log}$ & $\delta<1.25$ & $\delta<1.25^{2}$ & $\delta<1.25^{3}$\\
        \midrule
        \multirow{4}{*}{RepLKNet-B} & \multicolumn{3}{l}{Frozen} &  0 & 12.5 &  0.128 & 0.938 & 4.908 & 0.200 & 0.850 & 0.953 & 0.981\\ % gemini: b_sfz_s90000
        & \multicolumn{3}{l}{Full Fine-Tuned} &  78.8 & 12.5 &  0.092  &   0.774  &   4.355  &   0.175  &   0.911  &   0.966  &   0.982 \\ % result of rep0van_s80000
        % \cmidrule(r){2-12}
        & \multicolumn{3}{l}{Adapter (0.0625)} & 8.15 & 12.5 & \textbf{0.092} & \textbf{0.686} & \textbf{4.207} & \textbf{0.170} & 0.910 & \textbf{0.968} & \textbf{0.984}\\ % cbp625+++_s2700
        &\multicolumn{3}{l}{Adapter (0.25)} & 21.2 & 12.5 &  \textbf{0.090}  &   \textbf{0.666}  &   \textbf{4.175}  &  \textbf{0.168}  & \textbf{0.912}  &  \textbf{0.969}  &  \textbf{0.984}\\ % result of clcbfte6+_s150000
        \midrule
        \multirow{4}{*}{RepLKNet-L} &\multicolumn{3}{l}{Frozen}& 0 & 28.2 & 0.129 & 0.938 & 4.937 & 0.201 & 0.846 & 0.952 & 0.980 \\ % gemini: rep l special fz s72000
        &\multicolumn{3}{l}{Full Fine-Tuned}& 171 & 28.2 & 0.089 & 0.734 & 4.306 &  0.169 & 0.917 & 0.968&  0.983\\ % repl0van_s108000
        &\multicolumn{3}{l}{Adapter (0.0625)}& 18.1 & 28.2 & 0.090  &   \textbf{0.666} &  \textbf{4.146} &  \textbf{0.168} &  0.915  &   \textbf{0.969}  &  \textbf{0.985} \\ % gemini_l_625_t2_s72000; load and eval
        &\multicolumn{3}{l}{Adapter (0.25)}& 47.6 & 28.2 & \textbf{0.088}  &  \textbf{0.649}  &  \textbf{4.105}  &   \textbf{0.167}  &  \textbf{0.917}  &  \textbf{0.968}  &  \textbf{0.984}\\ % l_dprp3_s72000; load and eval
        \bottomrule    
    \end{tabular}}
    \caption{\textbf{Encoder Adapters are Effective for Domain Adaptation.} Our method achieves better depth estimation accuracy with fewer tuned parameters. Numbers in brackets after \textit{Adapter} indicate the bottleneck ratio.}
\label{tab:fullft}
\end{table*}

\subsection{Datasets}
Our method comprises two stages. The domain adaptation stage is trained and evaluated on the KITTI dataset, as it contains a substantial number of scenes that adhere to the static-scene assumption. Subsequently, the scene adaptation stage is built upon the parameters acquired from the domain adaptation stage on KITTI. This stage is then evaluated on more challenging datasets, including CityScapes and DDAD.

% 两个权威的benchmark
\subsubsection{KITTI}
% KITTI 比较符合静态假设，引用文献
The KITTI dataset \cite{geiger2012we} serves as the standard benchmark for evaluating self-supervised monocular depth estimation methods. We adhere to the established training protocols \cite{eigen2014depth} and utilize the data pre-processing approach introduced by \cite{Zhou_Brown_Snavely_Lowe_2017}, yielding 39,810 monocular triplets for training, 4,424 for validation, and 697 for testing. 
\subsubsection{CityScapes}
The CityScapes dataset \cite{cordts2016cityscapes} is notably more challenging due to its inclusion of numerous dynamic scenes with multiple moving objects \cite{casser2019unsupervised}. While fewer results are reported compared to KITTI, CityScapes serves as a prominent benchmark for those studies that focus on developing algorithms to handle dynamic objects \cite{lee2021learning,li2021unsupervised,feng2022disentangling,hui2022rm}. Following the setup of previous work \cite{feng2022disentangling}, we train on 58,355 and evaluate 1,525 images.

\subsubsection{DDAD}
DDAD is a more recent autonomous driving dataset \cite{guizilini20203d}. It is challenging owing to its extended depth range of up to 200 meters and inclusion of moving objects \cite{guizilini2022multi}. We follow the official DGP codebase of DDAD dataset to load images, and use 12,350 monocular triplets for train and 3,850 for evaluation.

\subsection{Evaluation Details} 
\label{sec:evaldetails}
As self-supervised learning predicts relative depth, we adhere to the established practice of scaling it before conducting evaluations \cite{monodepth2}. We use standard depth assessment metrics \cite{eigen2015predicting}, encompassing absolute and squared relative errors (AbsRel and SqRel), root mean squared error (RMSE), root mean squared log error (RMSElog), and accuracy within a threshold ($\delta$).

PPEA-Depth adopts the well-established multi-frame inference and teacher-student distillation training scheme \cite{watson2021temporal,feng2022disentangling,guizilini2022multi,bangunharcana2023dualrefine}. The main network contains a cost volume construction process, using both the current frame $I_{t}$ and its preceding frame $I_{t-1}$ to predict depth $D_t$. The teacher network does not involve cost volume generation and only takes the current frame during inference. Albeit for the difference in inner structure, the teacher and student share the same adapter design and training paradigm. Please refer to the supplementary for more details.

We carry out experiments using two variations of RepLKNet, each with different scales of parameter counts: RepLKNet-B and RepLKNet-L \cite{replknet}. In line with the approach taken by \citet{houlsby2019parameter,he2021towards,yang2023aim}, all adapter weights are initialized to zero to ensure stable training.

\subsection{Stage 1: Domain Adaptation}
Here we demonstrate the effectiveness of our adapter-based tuning strategy in the domain adaptation stage. We compare our method with two baselines on the KITTI dataset, all using the same pre-trained RepLKNet as the depth encoder. The first baseline (Frozen) freezes the depth encoder and only trains the depth decoder. The second baseline (Full Fine-Tuned) tunes all the parameters of the depth encoder and decoder. The goal of our domain adaptation stage is to add a few tunable parameters to the first baseline and close the gap between it and the full fine-tuning method. 

Encoder adapters project input features to a lower dimensional space and then project them back. Bottleneck ratio is the ratio between the input and the intermediate feature channels and directly influences the number of adapter parameters. We control the number of adapter parameters by setting different bottleneck ratios. As shown in Table \ref{tab:fullft}, for RepLKNet-B, the frozen encoder with our adapters achieves comparable performance with fine-tuning the entire backbone, while using 90\% fewer parameters than it. Tuning 21.2M encoder adapter parameters surpasses the performance of full fine-tuned RepLKNet-B. Experimental results for RepLKNet-L are similar. Our method reduces the number of tunable parameters by up to 90\% to achieve comparable performance with the full fine-tuned RepLKNet-L.

Our domain adaptation stage can preserve and utilize the generalized patterns in the ImageNet-pre-trained model by freezing the encoder backbone. 
It's worth noting that adapter tuning leads to significantly lower SqRel and RMSE values. This observation suggests that the generalized patterns retained through our adapter tuning strategy are beneficial for reducing extreme depth estimation errors.

\subsection{Stage 2: Scene Adaptation}
\begin{figure}[t!]
    \centering
    \includegraphics[width=\linewidth]{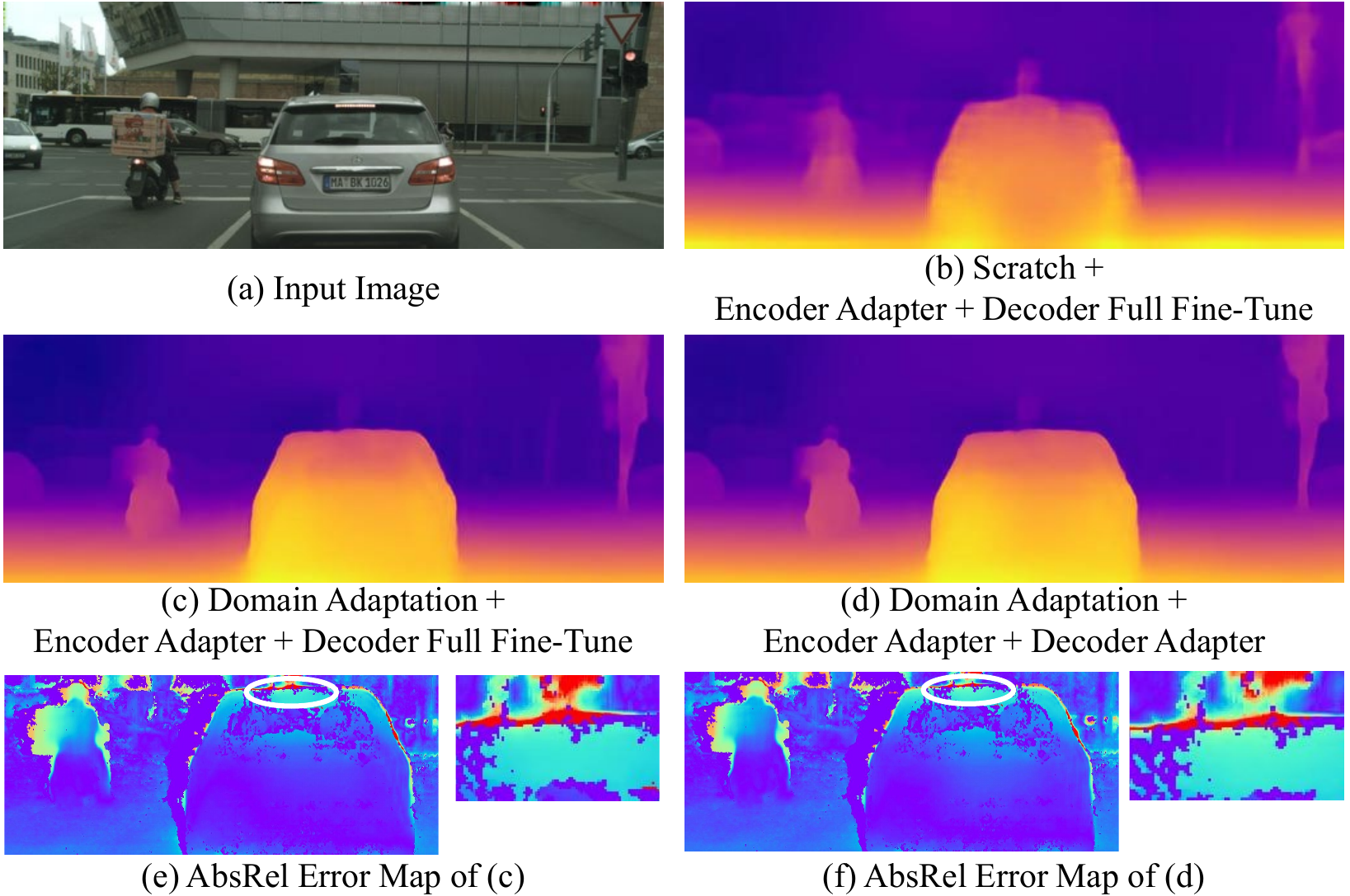}
    \caption{\textbf{Comparisons of Different Training Strategies on CityScapes}. Training from domain adaptation yields better depth estimates on vehicles and cyclists compared to training from scratch. Tuning the decoder adapter demonstrates improved depth estimates in the upper portion of the car compared to the full fine-tuned decoder.}
    \label{fig:scene_adapt}
\end{figure}
\begin{table*}[h]
    \centering
    \small
    \scalebox{1.0}{
    \begin{tabular}{cccccccccc}
        \toprule
        {\multirow{2}{*}{Train}} & \multicolumn{2}{c}{Tuning Strategy} & \multicolumn{4}{c}{\emph{Errors}$\downarrow$} & \multicolumn {3}{c}{\emph{Accuracy}$\uparrow$}\\
        \cmidrule(r){2-3} \cmidrule(r){4-7} \cmidrule(r){8-10} 
         {From}& {Encoder} & {Decoder}  & AbsRel & SqRel & RMSE & RMSE$_\mathrm{log}$ & $\delta<1.25$ & $\delta<1.25^{2}$ & $\delta<1.25^{3}$\\
        \midrule
        \multirow{2}{*}{Scratch} & Full Fine-Tuned & Full Fine-Tuned & 0.130 & 1.717 & 6.448 & 0.184 & 0.857 & 0.958 &  0.984\\  % b_cs_scratch_full_s36000 \\
         & Adapter  & Full FT. & 0.130 & 1.473 & 6.735 & 0.179 & 0.859  & 0.964 & 0.988 \\ % result of b_cs_scratch_s12000
        \cmidrule(r){1-10}
        & Original Weights & Original Weights & 0.138 &  1.319 & 7.211 & 0.198 & 0.819 & 0.957 & 0.987\\
        {Domain}& Full FT. & Full FT. & 0.116 & 1.120 & 6.193 & 0.168 & 0.873 & 0.969 & 0.991 \\ % fullft_s2000 
        {Adaptation}& Adapter & Full FT. & \textbf{0.103} & \textbf{0.962} & \textbf{5.716} & \textbf{0.155} & \textbf{0.897} & \textbf{0.976} & \textbf{0.992}\\ % k2cswodyn_s6000
        & Adapter & Adapter & \textbf{0.100} & \textbf{0.976} & \textbf{5.673} & \textbf{0.152} & \textbf{0.904} & \textbf{0.977} & \textbf{0.992}\\ % k2cs_dadpt_s17000
        % Stage & Adapters & Adapters & 0.104 & 1.160 & 5.939 & 0.156 & 0.900 &  0.975 & 0.991\\ %k2cs_dadpt_part_s10000
        \bottomrule    
    \end{tabular}}
    \caption{\textbf{Effectiveness of Our Scene Adaptation Strategy and Decoder Adapter}. We compare different training strategies on CityScapes \cite{cordts2016cityscapes}. Our strategy tunes based on the domain adaptation stage, yielding better estimated depth than training from scratch. The last two rows indicate tuning decoder adapter performs better than tuning the whole U-Net decoder.}
\label{tab:scene_adapt}
\end{table*}

\begin{table*}[h]
    \centering
    \small
    \scalebox{1.0}{
    \begin{tabular}{ccccccccccc}
        \toprule
        \multicolumn{2}{c}{\multirow{2}{*}{Percentage of}} &  \multicolumn{2}{c}{Tuning Strategy} & \multicolumn{4}{c}{\emph{Errors}$\downarrow$} & \multicolumn {3}{c}{\emph{Accuracy}$\uparrow$}\\
        \cmidrule(r){3-4} \cmidrule(r){5-8} \cmidrule(r){9-11} 
        \multicolumn{2}{c}{Training Data} & {Encoder} & {Decoder}  & AbsRel & SqRel & RMSE & RMSE$_\mathrm{log}$ & $\delta<1.25$ & $\delta<1.25^{2}$ & $\delta<1.25^{3}$\\
        \midrule
        \multicolumn{2}{c}{3\%} & Adapter & Adapter & 0.109 & 1.177 & 6.180 &  0.162 & 0.889 & 0.973 & 0.990 \\% pct_p3_s600
        \multicolumn{2}{c}{5\%} & Adapter & Adapter & 0.107 & 1.137 & 6.081 &  0.160 & 0.892 & 0.973 & 0.991 \\% pct_p5_s1600
        
        % \multicolumn{2}{c}{2.5\%} & Adapter & Adapter & 0.111 & 1.202 & 6.253 &  0.164 & 0.887 & 0.972 & 0.990 \\% pct_p2p5_s960
        \multicolumn{2}{c}{10\%} & Adapter & Adapter & 0.105 & 1.134 & 5.997 & 0.158 & 0.896 & 0.974 & 0.991\\ % pct_p10_s1800
        % \multicolumn{2}{c}{15\%} & Adapter & Adapter & \\ % pct_p10_s1800
        
        \multicolumn{2}{c}{25\%} & Adapter & Adapter  & 0.102 &  1.104 & 5.844& 0.154 & 0.902 & 0.975 &0.991\\ %k2cs_dadpt_part3_s7500
        
        \bottomrule    
    \end{tabular}}
    \caption{\textbf{Our Scene Adaptation and Adapters Enhance Data Efficiency}. Our scene adaptation strategy outperforms training from scratch and full fine-tuning from domain adaptation by a large margin even when utilizing only 3\% of the training data from CityScapes \cite{cordts2016cityscapes}.}
\label{tab:scene_adapt_sup}
\end{table*}

We verify the efficacy of our second training stage: the scene adaptation stage and its decoder adapter. We consistently select RepLKNet-B as the depth encoder and set the adapter bottleneck ratio to 0.25. We present a comparison on CityScapes under varying training strategies in Table \ref{tab:scene_adapt}.

Our proposed paradigm trains based on the weights learned in the first training stage on KITTI. Quantitative results in Table \ref{tab:scene_adapt} and qualitative results in Fig. \ref{fig:scene_adapt}(b-d) show that tuning on KITTI before training on CityScapes significantly outperforms the method of training from scratch.  There are a large number of dynamic objects in CityScapes, making training from scratch on it challenging because moving objects disrupt the pixel correspondences computed by Eqn. (\ref{formula:warp}). This may mislead the loss, resulting in updates to the network's parameters in incorrect directions.

We directly load the model trained from Stage 1 on KITTI and evaluate it on CityScapes (the third row of Table \ref{tab:scene_adapt}). Its outcomes are unsatisfactory, signifying that there exist notable distinctions between the two datasets.
We also assess another baseline: starting from weights learned on KITTI and subsequently full fine-tuning both the encoder and decoder on CityScapes (the fourth row of Table \ref{tab:scene_adapt}). It outperforms methods that train from scratch but evidently falls short in comparison to the adapter-tuning approach that also starts from the weights learned in the domain adaptation stage. This observation suggests that leveraging knowledge from earlier training phases benefits dynamic object handling, but directly fine-tuning all parameters disrupts previously acquired patterns. 

As evident from the last two rows in Table \ref{tab:scene_adapt}, fine-tuning the decoder adapter with only 0.185M parameters yields superior outcomes compared to fine-tuning the entire U-Net decoder. As illustrated in Fig. \ref{fig:scene_adapt} (e-f), tuning decoder adapter produces more precise depth estimations compared to the full fine-tuned decoder. Freezing both the U-Net encoder and decoder better preserves the depth priors acquired in previous stages and enhances the robustness towards errors in training losses caused by object motion. Our method, which merely tunes encoder adapters and the decoder adapter, offers a solution that strikes a balance between adapting the network to new datasets and conserving valuable depth perception patterns gained from preceding training phases.

% 数量
Our method also enhances data efficiency. We conducted Stage 2 training using randomly sampled subsets of 2.5\%, 10\%, and 25\% of the data from the CityScapes training set. The evaluation results for each subset are presented in Table \ref{tab:scene_adapt_sup}. Notably, tuning encoder adapters and decoder adapters with just around 3\% of the training data significantly surpasses the strategies of training from scratch and full fine-tuning the entire U-Net in Stage 2. Training with approximately 25\% of the data yields results comparable to tuning adapters on the entire training set, particularly for the accuracy metric $\delta<1.25$. This suggests that the adapter-based learning scheme can swiftly enhance the accuracy of depth estimation even with a limited amount of data.

\subsection{Depth Evaluation}
\begin{table*}[t!]
    \centering
    \scalebox{0.9}{
    \begin{tabular}{ccccccccccccc}
    \toprule
    \multicolumn{2}{c}{\multirow{2}{*}{Method}} & \multicolumn{1}{c}{\multirow{2}{*}{Test}} & \multicolumn{1}{c}{\multirow{2}{*}{S}} & \multicolumn{1}{c}{\multirow{2}{*}{W$\times$H}} &
    \multicolumn{4}{c}{\emph{Errors}$\downarrow$} & \multicolumn {3}{c}{\emph{Accuracy}$\uparrow$}\\
    \cmidrule(r){6-9} \cmidrule(r){10-12}
    \multicolumn{2}{c}{} &  Frames & {} & & AbsRel & SqRel & RMSE & RMSE$_\mathrm{log}$ & $\delta<1.25$ & $\delta<1.25^{2}$ & $\delta<1.25^{3}$\\
    \midrule
    \multicolumn{2}{l}{\citet{casser2019unsupervised}} & 1 &\CIRCLE &416$\times$128& 0.141 & 1.026 & 5.291 & 0.215 & 0.816 & 0.945 & 0.979\\ % Struct2Depth
    \multicolumn{2}{l}{\citet{bian2019unsupervised}} & 1 & &416$\times$128&  0.137 & 1.089 & 5.439 & 0.217 &  0.830 & 0.942 & 0.975\\
    \multicolumn{2}{l}{\citet{gordon2019depth}}  & 1 &\CIRCLE & 416$\times$128 & 0.128 & 0.959 & 5.230 & 0.212 & 0.845 & 0.947 & 0.976\\ %videos in the wild
    \multicolumn{2}{l}{\citet{monodepth2}}  & 1 & & 640$\times$192 & 0.115 & 0.903 & 4.863 & 0.193 & 0.877 & 0.959 & 0.981\\
    \multicolumn{2}{l}{\citet{lee2021learning}}  &1 & \CIRCLE& 832$\times$256  & 0.112 & 0.777 & 4.772 & 0.191 & 0.872 & 0.959 & 0.982\\ %InstaDM
    \multicolumn{2}{l}{\citet{guizilini20203d}}  & 1 & &640$\times$192  & 0.111 & 0.785 & 4.601 & 0.189 & 0.878 & 0.960 & 0.982\\ % Packet-SFM
    \multicolumn{2}{l}{\citet{hui2022rm}}  & 1 & & 640$\times$192 & 0.108 & 0.710 & 4.513 & 0.183 & 0.884 & 0.964 & 0.983 \\ % RM-Depth
    \multicolumn{2}{l}{\citet{Wang_Wang_Wang_Zhan_Wang_Lu_2021}}   & 1 & & 640$\times$192 &  0.109 & 0.779 & 4.641 & 0.186 & 0.883 & 0.962 & 0.982 \\
    \multicolumn{2}{l}{\citet{johnston2020self}}   & 1& & 640$\times$192 &  0.106 & 0.861 & 4.699 & 0.185 & 0.889 & 0.962 & 0.982\\
    
    \multicolumn{2}{l}{\citet{GuiziliniHLAG20}}   &1 &\CIRCLE & 640$\times$192 &  0.102 & {0.698} & {4.381} & 0.178 & 0.896 & 0.964 & {0.984}\\
    
    \multicolumn{2}{l}{\citet{wang2020self}}  & 2(-1,0) & & 640$\times$192 &  0.106 & 0.799 & 4.662 & 0.187 & {0.889} & {0.961} & {0.982}\\ 
    
    \multicolumn{2}{l}{\citet{watson2021temporal}}  & 2(-1,0) & & 640$\times$192 &  0.098 & 0.770 & 4.459 & 0.176 & {0.900} & {0.965} & {0.983}\\ 
    \multicolumn{2}{l}{\citet{feng2022disentangling}} & 2(-1,0) & \CIRCLE & 640$\times$192 &  {0.096} & 0.720 & 4.458 & {0.175} & 0.897 &  0.964 & {0.984} \\
    \multicolumn{2}{l}{\citet{guizilini2022multi}} & 2(-1,0)&  & 640$\times$192 & {0.090} & {0.661} & {4.149} & 0.175 & 0.905 & 0.967 & 0.984 \\
    \multicolumn{2}{l}{\citet{bangunharcana2023dualrefine}} & 2(-1,0) & & 640$\times$192 & \textbf{0.087} & 0.698 & 4.234 & 0.170 & 0.914 & 0.967 & 0.983 \\

    \midrule
    \multicolumn{2}{l}{{PPEA-Depth (RepLKNet-B)}}  &2(-1,0) & & 640$\times$192 &  {0.090}  &  {0.666}  &   {4.175}  &   \textbf{0.168}  &  {0.912}  &  \textbf{0.969}  &  \textbf{0.984}\\ % clcbfte6+_s15000 load and eval
    \multicolumn{2}{l}{{PPEA-Depth (RepLKNet-L)}}  & 2(-1,0)& & 640$\times$192 & {0.088}  &  \textbf{0.649}  &  \textbf{4.105}  &   \textbf{0.167}  &  \textbf{0.917}  &  \textbf{0.968}  &  \textbf{0.984}\\
    \bottomrule
\end{tabular}}
    \caption{\textbf{Depth Estimation Results on KITTI} Eigen Split \cite{eigen2015predicting}. S denotes a need for semantic information.} %. The best results are in \textbf{bold} while the second best is \underline{underlined}.}
\label{tab:sota_k}
\end{table*}

\begin{table*}[t!]
    \centering
    \scalebox{0.9}{
    \begin{tabular}{ccccccccccccc}
    \toprule
    \multicolumn{2}{c}{\multirow{2}{*}{Method}} & \multicolumn{1}{c}{\multirow{2}{*}{Test}} & \multicolumn{1}{c}{\multirow{2}{*}{S}} & \multicolumn{1}{c}{\multirow{2}{*}{W$\times$H}} &
    \multicolumn{4}{c}{\emph{Errors}$\downarrow$} & \multicolumn {3}{c}{\emph{Accuracy}$\uparrow$}\\
    \cmidrule(r){6-9} \cmidrule(r){10-12}
    \multicolumn{2}{c}{} & Frames & & & AbsRel & SqRel & RMSE & RMSE$_\mathrm{log}$ & $\delta<1.25$ & $\delta<1.25^{2}$ & $\delta<1.25^{3}$\\
    \midrule
    \multicolumn{2}{l}{\citet{casser2019unsupervised}} & 1 & \CIRCLE & 416$\times$128 & 0.145 & 1.737 & 7.280 & 0.205 & 0.813 & 0.942 & 0.976\\
    \multicolumn{2}{l}{\citet{monodepth2}} & 1& & 416$\times$128 & 0.129 & 1.569 & 6.876 & 0.187 & 0.849 & 0.957 & 0.983\\
    \multicolumn{2}{l}{\citet{gordon2019depth}} &1 & \CIRCLE& 416$\times$128 & 0.127 & 1.330 & 6.960 & 0.195 & 0.830 & 0.947 & 0.981\\
    \multicolumn{2}{l}{\citet{li2021unsupervised}} &1 &  & 416$\times$128 & 0.119 & 1.290 & 6.980 & 0.190 & 0.846 & 0.952 & 0.982\\
    \multicolumn{2}{l}{\citet{lee2021learning}} & 1 & \CIRCLE & 832$\times$256 & 0.111 & 1.158 & 6.437 & 0.182 & 0.868 & 0.961 & 0.983\\
    \multicolumn{2}{l}{\citet{watson2021temporal}} & 2(-1, 0) & & 416$\times$128 & 0.114 & 1.193 & 6.223 & 0.170 & 0.875 & 0.967 & 0.989\\
    \multicolumn{2}{l}{\citet{feng2022disentangling}} & 2(-1, 0) & \CIRCLE & 416$\times$128 & 0.103 & 1.000 & 5.867 & 0.157 & 0.895 &  0.974 & 0.991 \\
    \multicolumn{2}{l}{\citet{hui2022rm}} & 1 & & 416$\times$128 & 0.100 & \textbf{0.839} & 5.774 & 0.154 & 0.895 & 0.976 & \textbf{0.993} \\
    
    \midrule    
    \multicolumn{2}{l}{{PPEA-Depth (RepLKNet-B)}} & 1 & &416$\times$128 & \textbf{0.099} &  1.115  &  5.995  &  0.155  &  \textbf{0.905}  &   \textbf{0.976} &  0.991\\ %k2cs_dadpt_s11000
    \multicolumn{2}{l}{{PPEA-Depth (RepLKNet-B)}} & 2(-1, 0) & &416$\times$128 & \textbf{0.100} & 0.976 & \textbf{5.673} & \textbf{0.152} & \textbf{0.904} & \textbf{0.977} & 0.992\\
    % \multicolumn{2}{l}{{PPEA-Depth (RepLKNet-L)}} & 2(-1, 0) & &416$\times$128 & 0.103 & 1.053 & \textbf{5.709} & 0.157 & \textbf{0.900} & 0.973& 0.990\\
    \bottomrule
\end{tabular}}
    \caption{\textbf{Depth Estimation Results on CityScapes} \cite{cordts2016cityscapes}. S denotes a need for semantic information.} 
\label{tab:sota_cs}
\end{table*}

Table \ref{tab:sota_k} and Table \ref{tab:sota_cs} provide comprehensive comparisons between our method and state-of-the-art models on the two widely recognized benchmarks for self-supervised depth estimation: KITTI \cite{geiger2012we} and CityScapes \cite{cordts2016cityscapes}.
As specified in section \ref{sec:evaldetails}, PPEA-Depth incorporates both a teacher and a student network, following ManyDepth \cite{watson2021temporal}. The teacher network utilizes a single frame during inference, whereas the student network employs two frames (preceding and current). In Table \ref{tab:sota_k} and \ref{tab:sota_cs}, we present the results of the teacher and student networks by indicating the number of frames in the column of \textit{Test Frames} as 1 or 2.

Our model outperforms most previous models on both benchmarks. Specifically, our method demonstrates a notable enhancement in the accuracy metric $\delta < 1.25$, signifying a high degree of accurate inliers. In contrast to prior state-of-the-art models that mainly focus on improving the self-supervised monocular depth estimation methodology itself – such as designing intricate sub-modules for iterative refinement of estimated depth and pose \cite{bangunharcana2023dualrefine}, enhancing cost volume generation with transformers \cite{guizilini2022multi}, or incorporating a motion field prediction network \cite{hui2022rm} – our approach centers on introducing a more effective strategy for the training process. We aim to contribute by designing structures that leverage the generalized patterns in robust pre-trained models and creating more sensible learning paradigms to address challenging scenarios in self-supervised depth estimation.

\subsection{Generalization Ability}
To substantiate the generalization capability of our method, we extend our evaluation beyond the two standard benchmarks, and assess the performance of PPEA-Depth on a more recent dataset, DDAD \cite{guizilini20203d}. 
As detailed in Section \ref{sec:evaldetails}, the student network involves a cost volume construction process, which explicitly incorporates the depth range of the dataset. In the domain adaptation stage on KITTI, this range is set to 0-100m, which is not compatible with the DDAD dataset with a depth range of 0-200m. Therefore, we evaluate the teacher network on DDAD. 

Table \ref{tab:sota_d} compares our method with state-of-the-art models on DDAD. Using only a single frame during testing, our method outperforms previous models, showing the potent generalization capability of our adapters and the potential of transfer learning across varied datasets via adapter tuning with a core model. In this manner, we only need to maintain and incorporate a modest count of dataset-specific parameters, in addition to the core model, for each featured scene.

\begin{table}[t!]
    \centering
    % \small
    \scalebox{0.9}{
    \begin{tabular}{ccccccc}
    \toprule
    \multicolumn{2}{c}{\multirow{2}{*}{Method}} & \multirow{2}{*}{TF} & 
    \multicolumn{2}{c}{\emph{Errors}$\downarrow$} & \multicolumn {1}{c}{\emph{Accuracy}$\uparrow$}\\
    \cmidrule(r){4-6} 
    \multicolumn{2}{c}{} &  & AbsRel & SqRel & $\delta<1.25$ \\
    \midrule
    %RMSE$_\mathrm{log}$
    %\multicolumn{1}{c}{\multirow{2}{*}{W$\times$H}} &
    \multicolumn{2}{l}{\citet{guizilini20203d}} & 1 & 0.162 & 3.917   & 0.823 \\
    \multicolumn{2}{l}{\citet{guizilini2022multi}}& 2 & 0.135 & 2.953  & 0.836\\
    
    \midrule    
    \multicolumn{2}{l}{{PPEA-Depth (B)}} &1& \textbf{0.134} & \textbf{2.809}  & \textbf{0.836} \\ % gemini: k2ddad_b_mres_t2_s5600 load and val on server10
    \multicolumn{2}{l}{{PPEA-Depth (L)}} &1& \textbf{0.130} &  \textbf{2.695}  & \textbf{0.846}  \\ % gemini: k2ddad_l_mres_2_dc_t2_s22400
    \bottomrule
\end{tabular}}
    \caption{\textbf{Depth Estimation Results on DDAD } \cite{guizilini20203d} (WxH = 640x384) . TF represents Test Frames.} 
    
\label{tab:sota_d}
\end{table}

%% file: data/5conclusion.tex
\section{Conclusion}
In this paper, we introduce PPEA-Depth, a novel framework designed to enable the progressive transfer of pre-trained image models into the realm of self-supervised depth estimation. This transfer is orchestrated through the utilization of encoder and decoder adapters. Initially, the pre-trained model is tailored to accommodate depth perception using datasets primarily aligned with the static-scene assumption of self-supervised depth estimation methodology. Subsequently, it is further adapted to more challenging datasets involving a large number of moving objects. 

PPEA-Depth achieves state-of-the-art results on the KITTI and CityScapes, while also demonstrating its robust transferability on the DDAD dataset. Our method is promising to transfer to other tasks with loose-constrained loss, boosting network robustness towards loss errors.

%% file: data/a0exp.tex
\section{Supplementary Experiments}
\subsection{Robustness to Dynamic Scenes}
In our method, most network parameters are frozen and unaffected by the erroneous loss caused by object motion. Evidence is presented in Table \ref{tab:scene_adapt}, where tuning adapters (the last row) outperforms tuning the entire network (the third to last row) on CityScapes, a dataset with prevalent moving objects.

We conduct further experiments to validate the efficacy of our method by excluding areas of movable objects, such as cars and pedestrians, from the loss computation during training. Only the result of tuning the entire network is improved but our method cannot be enhanced (see Tab.~\ref{tab:dyn}). This shows the robustness of our approach to dynamic objects. Furthermore, even when excluding dynamic objects, tuning the entire network is still worse than our method, emphasizing the benefits of the adapter-tuning strategy.

\begin{table}[ht]
    \centering
    \scalebox{0.8}{
    \begin{tabular}{ccccc}
    \toprule
    Tuning Strategy & Dynamic Objects & AbsRel & SqRel & $\delta<1.25$\\
    \midrule
    \multirow{2}{*}{Full Fine-Tune} &Included & 0.116 & 1.120 & 0.873\\
    & Excluded & 0.106 & 1.000 & 0.890 \\% fullft_stmask_s3000 0.103 &  0.995 &  0.899\\ % fullftreb_womask_s6000% 0.106 & 1.05 & 0.896 \\ %
    \midrule
    Adapters & Included & 0.100 & 0.976 & 0.904 \\
    (Ours) &  Excluded & 0.102 & 1.113 & 0.904\\ % ed_adpt_stmask_s10000
    
    \bottomrule
\end{tabular}}
    \caption{PPEA-Depth's Robustness to Dynamic Objects.} 
\label{tab:dyn}
\end{table}

\subsection{Improvement with encoder adapters arises from a more complex model?}
Supplement to Table \ref{tab:fullft}, we did another experiment on Stage 1. We employ RepLKNet-B as the depth encoder and tune all the parameters of both the base model and encoder adapters. There is no significant performance gain (see Table~\ref{tab:s1}). This proves that our PEFT scheme is the main reason for the improvement shown in Table \ref{tab:fullft} and our approach can better exploit the pre-trained encoder than tuning all parameters.
%can efficiently summarize network updates with fewer parameters, which is beneficial for exploiting pre-trained models.
\begin{table}[ht]
    \centering
    \scalebox{0.8}{
    \begin{tabular}{cccccc}
    \toprule
    Encoder & Decoder & Encoder Adapters & AbsRel & SqRel & $\delta<1.25$\\
    \midrule
    % \CIRCLE & & 0.092 & 0.774 & 0.911\\ 
    Tuning & Tuning & Tuning & 0.093 & 0.703 & 0.911 \\
    % & \CIRCLE & 0.090 & 0.666 & 0.912 \\
    \bottomrule
\end{tabular}}
\caption{Tuning both encoder adapter and entire network is worse than freezing encoder and tuning encoder adapters.} 
\label{tab:s1}
\end{table}

\subsection{Comparison to Linear Probing}
PPEA-Depth can tune each layer in the network by adding adapters. It is more flexible than linear probing, where only the last linear layer can be tuned.
We've made a comparison with a strategy similar to classifier adjustment, where we freeze the encoder and only tune the decoder \textit{in Table \ref{tab:fullft} (Row 1)} for Stage 1. Results of this strategy for Stage 2 are also worse than our approach (see Table~\ref{tab:lp}), showing the necessity of encoder adapters for domain shift in Stage 2.
\begin{table*}[ht]
    \centering
    \scalebox{1.0}{
    \begin{tabular}{ccccccc}
    \toprule
    \multirow{2}{*}{Encoder} & \multirow{2}{*}{Encoder} & \multirow{2}{*}{Decoder} & \multirow{2}{*}{Decoder} & \multicolumn{2}{c}{\emph{Errors}$\downarrow$} & \multicolumn{1}{c}{\emph{Accuracy}$\uparrow$}\\
    \cmidrule(r){5-6} \cmidrule(r){7-7}
    & Adapters & & Adapter & AbsRel & SqRel & $\delta<1.25$\\
    \midrule
    Frozen & Frozen & Tuning & Not Used & 0.114 & 1.215 & 0.878\\ % lps2_s6000
   Frozen & Frozen & Frozen & Tuning & 0.107 & 1.004 & 0.889 \\
    \bottomrule
\end{tabular}}
\caption{Results of \textit{Classifier Adjustment} for Stage 2.} 
\label{tab:lp}
\end{table*}

%% file: data/a1ablations.tex
\section{Ablations for Adapter Design}
We conduct experiments to assess various adapter designs using RepLKNet-B. All the reported results are from the student network.
\subsection{Encoder Adapter}
\begin{table*}[ht]
    \centering
    \scalebox{0.9}{
    \begin{tabular}{ccccccccc}
        \toprule
       \multicolumn{1}{l}{\multirow{2}{*}{Adapter}} & \multirow{2}{*}{Params} & \multicolumn{4}{c}{\emph{Errors}$\downarrow$} & \multicolumn {3}{c}{\emph{Accuracy}$\uparrow$}\\
        \cmidrule(r){3-6} \cmidrule(r){7-9}
        \multicolumn{1}{l}{Position}& (M) & AbsRel & SqRel & RMSE & RMSE$_\mathrm{log}$ & $\delta<1.25$ & $\delta<1.25^{2}$ & $\delta<1.25^{3}$\\
        \midrule
        \multicolumn{1}{l}{RepLKBlock} & 17.7 & 0.093 & 0.718 & 4.285 & 0.171 & 0.909 & 0.968 & 0.984\\ % repblkonly+_s12000
        \multicolumn{1}{l}{ConvFFN} & 3.78 & 0.100 & 0.769 & 4.442& 0.175 & 0.900 &  0.966 & 0.984 \\ % convffn_s33000
        \multicolumn{1}{l}{All} & 21.9 & {0.090}  &  {0.666}  &  {4.175}  &  {0.168}  & {0.912}  &  {0.969}  &  {0.984}\\ % result of clcbfte6+_s150000
        \bottomrule    
    \end{tabular}}
    \caption{\textbf{Comparison of Attaching Adapters to Different Types of Blocks.} Attaching adapters to more sophisticated blocks in backbone leads to better results.}
\label{tab: type}
\end{table*}
\begin{table*}[ht]
    \centering
    \scalebox{0.9}{
    \begin{tabular}{cccccccccc}
        \toprule
       \multicolumn{1}{c}{\multirow{2}{*}{Down}} & \multicolumn{1}{c}{\multirow{2}{*}{Up}} & \multirow{2}{*}{Params} & \multicolumn{4}{c}{\emph{Errors}$\downarrow$} & \multicolumn {3}{c}{\emph{Accuracy}$\uparrow$}\\
        \cmidrule(r){4-7} \cmidrule(r){8-10}
        \multicolumn{1}{l}{Projector} & \multicolumn{1}{l}{Projector} & (M) & AbsRel & SqRel & RMSE & RMSE$_\mathrm{log}$ & $\delta<1.25$ & $\delta<1.25^{2}$ & $\delta<1.25^{3}$\\
        \midrule
        \multicolumn{1}{c}{Linear} & \multicolumn{1}{c}{Linear} & 7.5 &  0.095  &   0.776  &   4.372  &   0.174  &   0.909  &   0.967  &   0.983\\ % slimv3_s63000
        \multicolumn{1}{c}{Conv} & \multicolumn{1}{c}{Linear} & 21.9 & \textbf{0.090} & \textbf{0.666}  &  \textbf{4.175}  &  \textbf{0.168}  & \textbf{0.912}  &  \textbf{0.969}  &  \textbf{0.984}\\ % result of clcbfte6+_s150000
        \multicolumn{1}{c}{Conv} & \multicolumn{1}{c}{Conv} & 35.4 & \textbf{0.090} & 0.671 & 4.203 & 0.169 & \textbf{0.912} & 0.968 & \textbf{0.984}\\ % result of blkc3_s57000
        \bottomrule    
    \end{tabular}}
    \caption{\textbf{Influence of Encoder Adapter Receptive Field.} Domain adaptation stage benefits from employing encoder adapters with down projector of larger receptive fields, while the improvement brought by up projector of larger receptive fields is not obvious.}
\label{tab:proj}
\end{table*}
\begin{figure*}[!h]
    \centering
    \includegraphics[width=\textwidth]{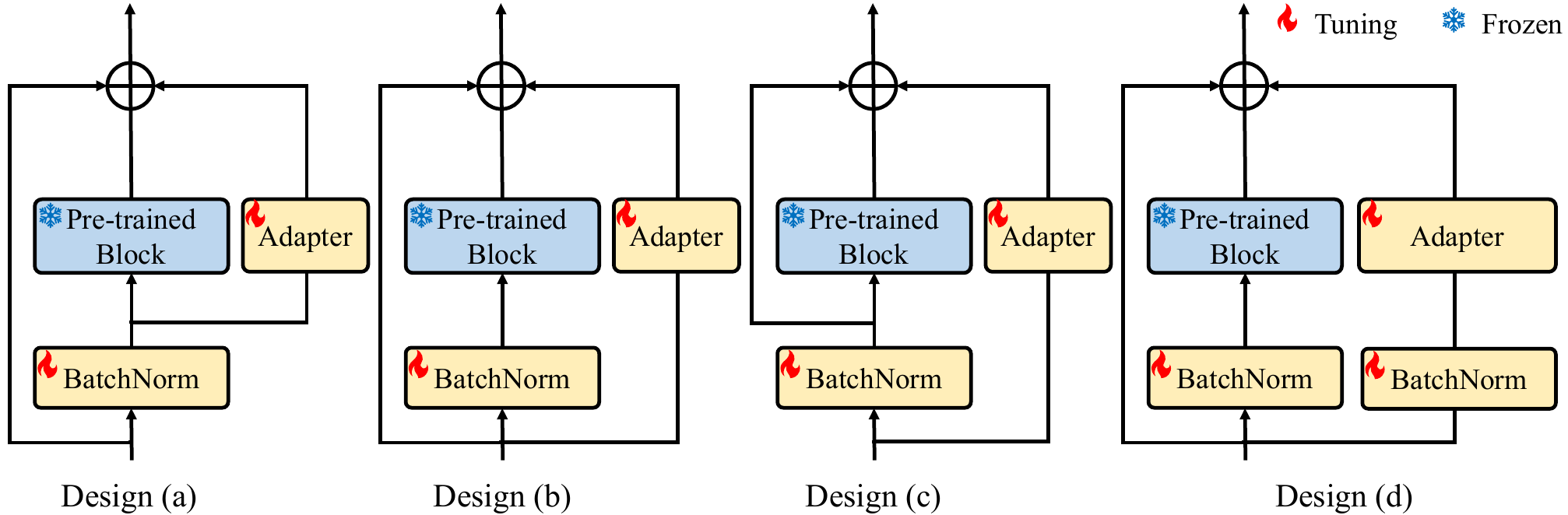}
    \caption{\textbf{Different Encoder Adapter Designs}. Design (a) is our final choice.}
    \label{fig: bnpos}
    % \vspace{-0.5cm}
\end{figure*}
\begin{table*}[!h]
    \centering
    \begin{tabular}{cccccccc}
        \toprule
        \multicolumn{1}{l}{\multirow{2}{*}{Design}} & \multicolumn{4}{c}{\emph{Errors}$\downarrow$} & \multicolumn {3}{c}{\emph{Accuracy}$\uparrow$}\\
        \cmidrule(r){2-5} \cmidrule(r){6-8}
        &  AbsRel & SqRel & RMSE & RMSE$_\mathrm{log}$ & $\delta<1.25$ & $\delta<1.25^{2}$ & $\delta<1.25^{3}$\\
        \midrule
        {a} & \textbf{0.090}  &   \textbf{0.666}  &   \textbf{4.175}  &  \textbf{0.168}  & \textbf{0.912}  &  \textbf{0.969}  &  \textbf{0.984}\\ %clcbfte6+_s15000
        {b} &  0.095 & 0.769 &  4.407 & 0.175 & 0.905 & 0.965 & 0.983\\ % bn_ablate_s51000
        {c} & 0.103 & 0.761 & 4.499 & 0.181 & 0.890 & 0.962 & 0.983\\ % bn_vc_s66000 
        {d} & 0.091 & 0.680 & 4.219 & 0.170 & 0.911 & 0.968 & 0.984\\ % new_adpt_bn_s72000
        \bottomrule    
    \end{tabular}
    \caption{\textbf{Comparison of Different BatchNorm and Skip Connection Designs} in Encoder Adapters.}
\label{tab: bn}
\end{table*}
\begin{table*}[!h]
    \centering
    \begin{tabular}{ccccccccc}
        \toprule
        {\multirow{2}{*}{Input}} & \multirow{2}{*}{Params} &
        \multicolumn{4}{c}{\emph{Errors}$\downarrow$} & \multicolumn {3}{c}{\emph{Accuracy}$\uparrow$} \\
        \cmidrule(r){3-6} \cmidrule(r){7-9}
        {Scales} & {(M)}  & AbsRel & SqRel & RMSE & RMSE$_\mathrm{log}$ & $\delta<1.25$ & $\delta<1.25^{2}$ & $\delta<1.25^{3}$\\
        \midrule
        {3} & 0.149 & 0.103 & 0.977 & 5.693 & 0.157 & 0.895 &  0.974 & 0.991\\%  dec_abl_s0_s6000
        {0, 3} & 0.185 & \textbf{0.100} & \textbf{0.976} & {5.673} & \textbf{0.152} & \textbf{0.904} & \textbf{0.977} & \textbf{0.992}\ \\ % result of v1_full_enc_s2300
        {0, 1, 2, 3} & 0.486 & 0.101 & 0.982 & \textbf{5.666} & \textbf{0.152} & \textbf{0.904} & \textbf{0.977} & \textbf{0.992}\\ % dec_4s_s17000
        \bottomrule    
    \end{tabular}
    \caption{\textbf{Comparison of Different Input Feature Levels} to Decoder Adapter.}
    \label{tab:dec_in}
\end{table*}
We compare different encoder adapter designs for the domain adaptation stage on the KITTI dataset.
\subsubsection{Type of Blocks to Attach Adapters}
Our method involves attaching encoder adapters to two types of blocks within the RepLKNet \cite{replknet} backbone: RepLKBlock and ConvFFN. We conducted an ablation study by adding adapters to only one type of these blocks. The results are presented in Table \ref{tab: type}. Evidently, the frozen encoder benefits more from adding adapters to RepLKBlock rather than ConvFFN. RepLKBlock, which leverages a large kernel size of up to 31x31, stands as the core innovation of RepLKNet. This outcome aligns with the intuition that adapters on more intricate blocks carry greater significance.

\subsubsection{Receptive Field of Projectors} 
Adapters project the input feature down to a lower dimension and then project it up. As depth estimation is a regression task, will encoder adapters benefit from a larger receptive field? We conducted experiments using different receptive fields for the adapters of RepLKBlock, which, as indicated by the results in Table \ref{tab: type}, holds greater importance than ConvFFN. We compared two types of projectors: linear projection and convolution with a kernel size of 3, while maintaining a bottleneck ratio of 0.25. The results are presented in Table \ref{tab:proj}. From the results, it can be inferred that all metrics benefit from a larger receptive field in the down projector, while the enhancement resulting from a wider receptive field in the up projector is not as apparent.

\subsubsection{BatchNorm and Skip Connections}
We experiment with different encoder adapter designs by varying the position of the BatchNorm module and skip connections (Fig. \ref{fig: bnpos}), while keeping the inner adapter design the same (bottleneck ratio is 0.25, down projector is a 3x3 convolution and up projector is linear). The evaluation results of different designs are in Table \ref{tab: bn}. 

The BatchNorm module before the pre-trained block is part of the original RepLKBlock and ConvFFN design (more details are shown in Fig. \ref{fig: replkdetail}). We design to let the adapter share the BatchNorm module with the original pre-trained block, and unfreeze the parameters of the BatchNorm during training. From the results in Table \ref{tab: bn}, it can be concluded that:
\begin{itemize}
    \item The final depth estimation accuracy benefit from the BatchNorm block before the encoder adapter. Removing the prefix BatchNorm before encoder adapters results in an obvious performance loss (Design (b)).
    \item The BatchNorm module before the pre-trained block is indispensable (Design (c)). As the parameters in the pre-trained blocks are frozen, the frozen parameters are matched with the output of the prefix BatchNorm, removing the prefix BatchNorm for the pre-trained block leads to degraded results. 
    \item Sharing the same prefix BatchNorm module between the pre-trained block and the encoder adapter (Design (d)) and training an extra BatchNorm for the encoder adapter (Design (a)) show no obvious difference in the final depth estimation errors and accuracy. The sharing strategy leads to a lower trainable parameter cost, so design (a) in Fig. \ref{fig: bnpos} is our choice. 
\end{itemize}

\subsubsection{Bottleneck Ratio.} 
The performance improves with more tunable parameters up to a bottleneck ratio of 0.25 in adapters, beyond which the benefits plateau (Table~\ref{tab:ratio}). 
\begin{table}[ht]
    \centering
    \scalebox{0.9}{
    \begin{tabular}{ccccc}
    \toprule
    {Ratio} & Tuning Params (M) & AbsRel & SqRel & $\delta<1.25$\\
    \midrule
    0.0375 & 6.4 & 0.098 & 0.779 & 0.902 \\
    0.0625 & 8.2 & 0.092 & 0.686 & 0.910 \\
    0.1 & 10.7 & 0.093 & 0.704 & 0.911\\
    % 0.2 & 17.7 & 0.092 & 0.696 & 0.910\\
    0.25 & 21.9 & 0.090 & 0.666 & 0.912\\
    0.5 & 38.7 & 0.090 & 0.674 & 0.912 \\ % b_p5_blk_s72000
    % 0.75 & 56.1 & 0.090 & 0.681 &  0.913 \\ % b_p75_blk_s99000
    1 & 73.6 & 0.090 & 0.659 & 0.911 \\ % b_p1_blk_s72000
    \bottomrule
\end{tabular}}
    \caption{\textbf{Effect of Adapter Bottleneck Ratio.}}

\label{tab:ratio}
\end{table}

\subsection{Decoder Adapter}
We compare different decoder adapter designs for the scene adaptation stage on the CityScapes dataset. We keep the encoder adapter design consistent (bottleneck ratio is 0.25, down projector is a 3x3 convolution, and up projector is linear) and use the same weights trained from the domain adaptation stage as initial model weights in the following experiments.

% \subsubsection{Bottleneck Ratio}
% \begin{table*}[t!]
%     \centering
%     % \aboverulesep=0ex % Solution part 1 of 3
%     % \belowrulesep=0ex
%     \begin{tabular}{ccccccccc}
%         \toprule
%         % {Method} & {RepLKBlk} & {ConvFFN} & {TransBlk} & {Params} & AbsRel$\downarrow$ & $\delta<1.25$\uparrow\\
%        \multicolumn{1}{l}{\multirow{2}{*}{Bottleneck}} & \multirow{2}{*}{Params} & \multicolumn{4}{c}{\emph{Errors}$\downarrow$} & \multicolumn {3}{c}{\emph{Accuracy}$\uparrow$}\\
%         \cmidrule(r){3-6} \cmidrule(r){7-9}
%         \multicolumn{1}{l}{Ratio} & (M) & AbsRel & SqRel & RMSE & RMSE$_\mathrm{log}$ & $\delta<1.25$ & $\delta<1.25^{2}$ & $\delta<1.25^{3}$\\
%         \midrule
%         \multicolumn{1}{l}{0.0625} & 0.053 & \\ % test_s17000
%         \multicolumn{1}{l}{0.1} & 0.079 \\ % interlace_s57500
%         \multicolumn{1}{l}{0.25} & 0.185 & 0.090 & 0.669 & 4.181 & 0.168 & 0.911 & 0.969 & 0.984\\ % 
%         % \multicolumn{1}{l}{0.3} & 24.7 &  &  &  &  &  &  & \\ % result of blkc3_s57000
        
%         \bottomrule    
%     \end{tabular}
%     \caption{\textbf{Effect of Different Bottleneck Ratio on Decoder Adapter}.}
% \label{tab: ratio}
% \end{table*}
\subsubsection{Input Scales}
The depth encoder generates feature maps at four levels ($F^{0}, F^{1}, F^{2}, F^{3}$), corresponding to $1/4$, $1/8$, $1/16$, and $1/32$ of the original image spatial shape.   Upsampling feature maps at different levels to the $1/4$ scale, we concatenate and feed them to the decoder adapters. We compare different input feature scales for the decoder adapter, and the results are in Table \ref{tab:dec_in}, suggesting that using the concatenation of the shallowest and deepest features as input is the optimal choice, considering both parameter count and performance.

%% file: data/a2latency.tex
\section{Inference Latency Introduced by Adapter}
As adapters are additional modules added to the network, we investigate the inference latencies introduced by the encoder and decoder adapters. We measure the total time required for the model to predict depth maps for 1,525 images in the CityScapes test dataset and calculate the average inference time per image. Our encoder backbone is RepLKNet-B, and we use a batch size of 12. The results are presented in Table \ref{tab:latency}. Thanks to the parallel data stream design, the inference time remains largely consistent whether adapters are used or not.
\begin{table}[ht]
    \centering
    \begin{tabular}{ccc}
        \toprule
        Encoder  & Decoder  & \multirow{2}{*}{Time (s)}\\
        Adapters &  Adapters &\\
        
        \midrule
        & & 0.037\\
        \CIRCLE & & 0.036\\
        \CIRCLE & \CIRCLE & 0.038\\
        \bottomrule
    \end{tabular}
    \caption{\textbf{Average Inference Time Per Image} on CityScapes.}
    \label{tab:latency}
\end{table}
% \begin{table*}[t!]
% \parbox{.45\linewidth}{
%     }
%     \hfill
%     \parbox{.45\linewidth}{
%     }
% \end{table*}

%% file: data/a3details.tex
\section{Supplementary Details}
\subsection{Overview of the Whole Framework}\label{sec: overview}
We follow the self-supervised depth estimation framework proposed by \citet{watson2021temporal}, whose depth network consists of a teacher network and a student network to improve estimated depths and better handle dynamic objects, and this design is also adopted in \citet{guizilini2022multi, feng2022disentangling}. The teacher depth network is trained jointly with the student, sharing the same pose predictions, and is discarded during evaluation.
\begin{figure*}[hb]
    \centering
    \includegraphics[width=0.7\textwidth]{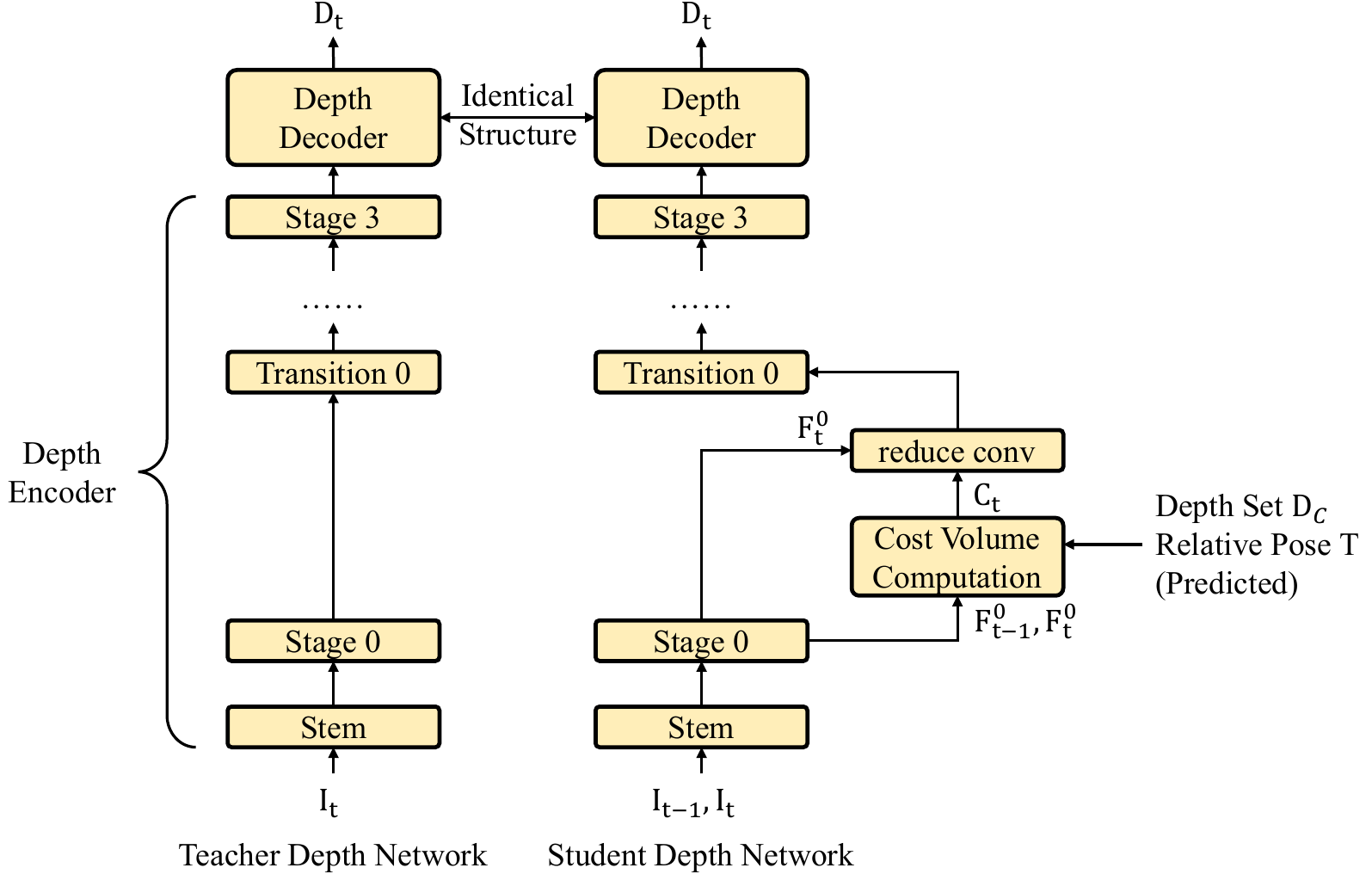}
    \caption{\textbf{Detailed Structure of Teacher and Student Depth Network.}}
    \label{fig: cv}
    % \vspace{-0.5cm}
\end{figure*}
Both the teacher and student networks are based on the U-Net architecture. They share identical designs for the depth network decoder but differ in their encoders. The teacher's input is a single image frame (the frame at timestamp $t$, denoted as $I_{t}$). This input, $I_{t}$, undergoes processing through the depth encoder and subsequently the depth decoder, leading to pixel-wise depth estimation.

As shown in Figure \ref{fig: cv}, the input of the student network includes two frames, $I_{t}$ and its previous frame $I_{t-1}$. The two frames are first separately extracted by the first stage of the encoder to generate first-level features $F^{0}_{t}, F^{0}_{t-1}$, which are both in the shape of $(B, C, H/4, W/4)$, where B is the batch size, C is the number of channels, H and W represent the spatial shape of $I_{t}$. Then a cost volume is built based on $F^{0}_{t}, F^{0}_{t-1}$, relative camera pose $T$ (which is currently predicted by the pose network), and a set of depth values $D_c$. 

Iterating over $D_c$, we project $F^{0}_{t-1}$ using $T$ and the iterated depth value $d_{i}$ to generate $F^{0}_{t-1\to t}$ according to the pixel correspondences mentioned in Equation (1) in the main paper. The L1 difference between $F^{0}_{t-1\to t}$ and $F^{0}_{t}$ is adopted to build the cost volume. The minimum and maximum depths ($d_{min}, d_{max}$) are initialized as 0.1m and 100m respectively, and are dynamically tuned in the learning process according to the strategy proposed in \citet{watson2021temporal}. The depth set $D_c$ is generated by uniformly sampling depth values in $[d_{min}, d_{max}]$ in logarithm space. 

The generated cost volume $C_{t}$ is in the shape of $(B, |D_{c}|, H/4, W/4)$, where $|D_{c}|$ represents the number of depths in set $D_{c}$. Then $C_{t}$ is concatenated with $F^{0}_{t}$ at the second dimension, and the concatenated feature is compressed to the shape of $(B, C, H/4, W/4)$ by a 3x3 convolution called \emph{reduce conv}. Then the output of the reduce conv is fed to the rest stages of the depth encoder and then to the depth decoder, and finally predicts a depth map $D_{t}$.

The cost volume design in the student network introduces $I_{t-1}$ and an iteration over all possible depths in the encoder, and exploits the relationship between two consequent frames to improve depth estimation. However, such a design exaggerates the depth estimation error in the dynamic object areas and tends to predict depths of such areas as infinity \cite{watson2021temporal}. 

\subsection{Teacher-Student Distillation Scheme}
To overcome the infinity-depth issue as mentioned above, a teacher-student training scheme is employed in our network as in previous works \cite{watson2021temporal,feng2022disentangling,guizilini2022multi}. 

For the dynamic object areas, the predictions from the student network are unreliable. Such unreliable area $M$ is computed by comparing the predicted depths from the teacher and student in a pixel-wise manner:$$
M=\max(\frac{D_{s}-D_{t}}{D_{t}}, \frac{D_{t}-D_{s}}{D_{s}}) > 1$$

During the training process, the teacher network is supervised by the image reprojection loss (as mentioned in Section 3.1 in the main paper). For the student network, the reliable area ($\neg{M}$) is also supervised by the image reprojection loss, while the unreliable area ($M$) is instead supervised by a depth consistency loss to enforce a knowledge distillation from the teacher. The depth consistency loss is the L1 difference of the predicted depths between the teacher and the student. In the depth consistency loss, gradients to depths predicted by the teacher are blocked, ensuring the distillation is unidirectional, i.e. only from teacher to student.

Different from the previous works \cite{watson2021temporal, feng2022disentangling,guizilini2022multi}, we do not freeze the teacher depth network and pose network to fine-tune in the last five epochs, since we do not observe a significant performance improvement with such technique on PPEA-Depth.

\subsection{Amount of Encoder Adapter Parameters}
All parameter counts mentioned in the main paper are based on the teacher network. Here, we provide supplementary details regarding the tunable parameter count for encoder and decoder adapters in the student network. Throughout all experiments, the settings for student encoder adapters (down projector, up projector, and bottleneck ratio) remain the same as those of the teacher network. The student network has more tunable parameters than the teacher because apart from the encoder adapters and the BatchNorm module, the reduce conv module (as detailed in Section \ref{sec: overview}) also receives updates. Refer to Table \ref{tab: param} for specific parameter details.
\begin{table}[]
    \centering
    \scalebox{1.0}{
    \begin{tabular}{ccccc}
        \toprule
        \multirow{2}{*}{Pre-trained} & \multirow{2}{*}{Adapter} & \multirow{2}{*}{Ratio} & \multicolumn{2}{c}{Tuning Params(M)}\\
        \cmidrule{4-5}
        {Backbone} & Type &  & Teacher & Student\\
        \midrule
        \multirow{3}{*}{RepLKNet-B} & Encoder & 0.0625 & 8.15 & 8.41 \\
        & Encoder & 0.25 & 21.9 & 22.2  \\
        & Decoder & 0.25 & 0.185 & 0.185\\
        \midrule
        \multirow{3}{*}{RepLKNet-L}& Encoder & 0.0625 & 18.1 & 18.6  \\
        & Encoder & 0.25 & 47.6 &  48.1\\
        & Decoder & 0.25 &  4.15 & 4.15 \\
        
        \bottomrule    
    \end{tabular}}
    \caption{\textbf{Details of Tunable Parameters} in Teacher and Student Depth Network Adapters.}
    \label{tab: param}
\end{table}

\subsection{RepLKNet}
In Section 3.3 of the main paper, we briefly introduce the main pipeline of RepLKNet-31B, which we adopt as the encoder backbone of both the teacher and student depth network. Here we introduce the detailed structure of the two key modules to which we attach encoder adapters, RepLKBlock and ConvFFN. For more details please refer to \citet{replknet}. 

\begin{figure*}[ht]
    \centering
    \includegraphics[width=0.7\textwidth]{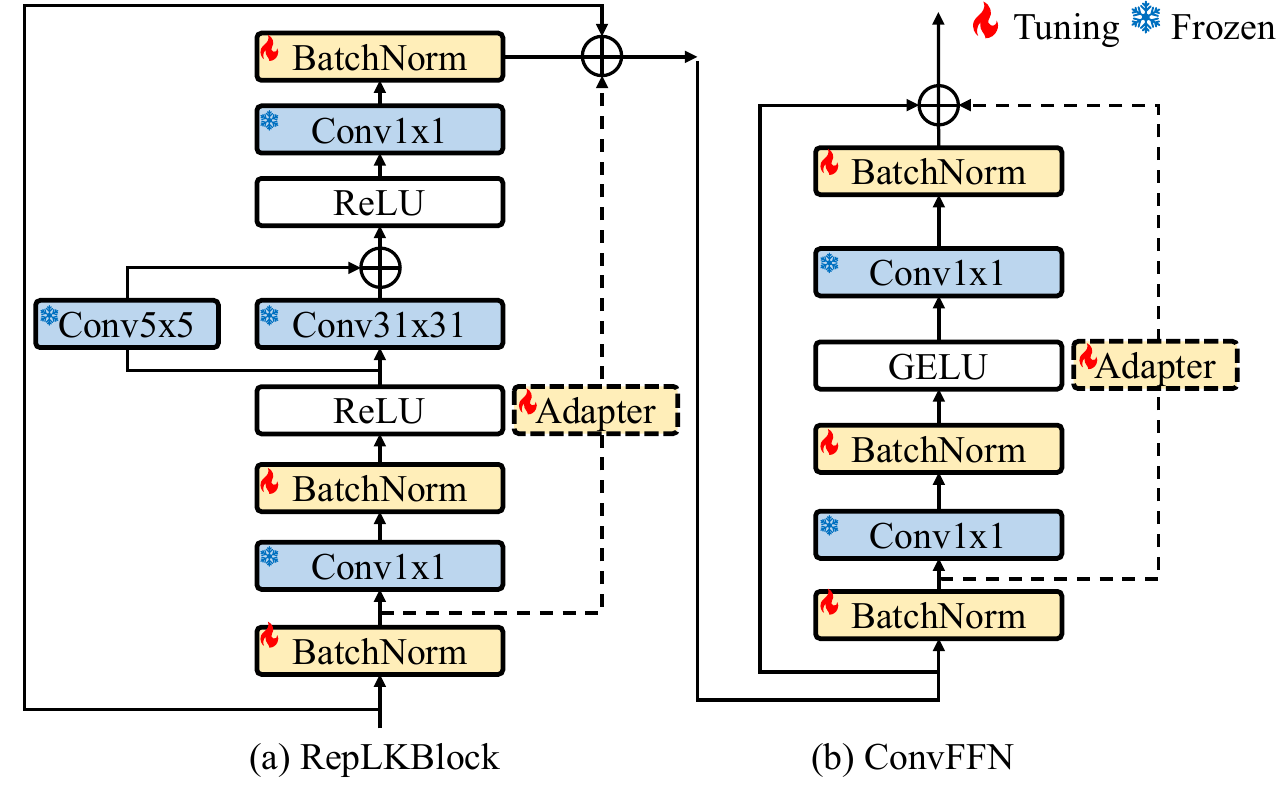}
    \caption{\textbf{Detailed Structure of RepLKBlock and ConvFFN}. The dashed line depicts the position of encoder adapters in PPEA-Depth.}
    \label{fig: replkdetail}
\end{figure*}

The detailed structures of RepLKBlock and ConvFFN are shown in Figure \ref{fig: replkdetail}. Our proposed encoder adapters share the same prefix BatchNorm module with RepLKBlock and ConvFFN (as in Figure \ref{fig: bnpos}). Since we adopt a different batch size during training compared with the RepLKNet pre-training process, all parameters of BatchNorm modules are not frozen. 

\subsection{Training Details}
PPEA-Depth is implemented with PyTorch \cite{paszke2017automatic} and is trained on GeForce RTX 3090 GPU. We use the Adam optimizer \cite{kingma2014adam}, with $\beta_{1}=0.9, \beta_{2}=0.999$.  At the domain adaptation stage (Stage 1), the batch size is set to 12. This stage trains for 25 epochs, and the initial learning rate is 1e-4. The learning rate decays to 1e-5 after 15 epochs and decays to 1e-6 after 20 epochs. During the scene adaptation stage (Stage 2), we utilize a batch size of 24 for training. This stage is conducted over 10 epochs, employing a learning rate of 1e-5 for CityScapes and 5e-5 for DDAD.

\subsection{Evaluation Metrics Details}
We evaluate our depth estimation results using the standard depth assessment metrics \cite{eigen2015predicting}, including Absolute Relative Error (\textit{AbsRel}), Squared Relative Error (\textit{SqRel}), Root Mean Squared Error ($\mathrm{RMSE}$), Root Mean Squared Log Error ($\mathrm{RMSE}_\mathrm{log}$), $\delta_{1}$, $\delta_{2}$, and $\delta_{3}$. The specific formulas to calculate these metrics are as follows:
\begin{equation}
    \begin{aligned}
        AbsRel &= \frac{1}{n}\sum_{i}{\frac{p_{i}-g_{i}}{g_{i}}}\\\\ 
        SqRel &= \frac{1}{n}\sum_{i}{\frac{(p_{i}-g_{i})^{2}}{g_{i}}} \\\\
        \mathrm{RMSE} &= \sqrt{\frac{1}{n}\sum_{i}{(p_{i}-g_{i})^{2}}}\\\\
        \mathrm{RMSE_{log}} &= \sqrt{\frac{1}{n}\sum_{i}{(\mathrm{log}p_{i}-\mathrm{log}g_{i})^{2}}}\\
    \end{aligned}
\end{equation}
and the values $\delta_{1}$, $\delta_{2}$, and $\delta_{3}$ represent the percentages of pixels where the condition $\mathrm{max}\ (p/q,\  q/p) < 1.25, 1.25^{2}, 1.25^{3}$ is satisfied. Here, $g$ denotes the ground truth depth, $p$ stands for the predicted depth, and $n$ represents the number of pixels.

%% file: data/a4qualitative.tex
\section{Supplementary Qualitative Results}
We provide more qualitative comparisons on CityScapes in the last two pages. The images are organized from left to right, showcasing the original image, the estimated depth obtained by full fine-tuning a U-Net from scratch, and the estimated depth produced by our PPEA-Depth approach. We also provide a qualitative video demo (\textit{demo.mp4} in the .zip file). Please check it out for a more dynamic representation of our approach's performance.
\begin{figure*}[ht]
    \centering
    \includegraphics[width=\textwidth]{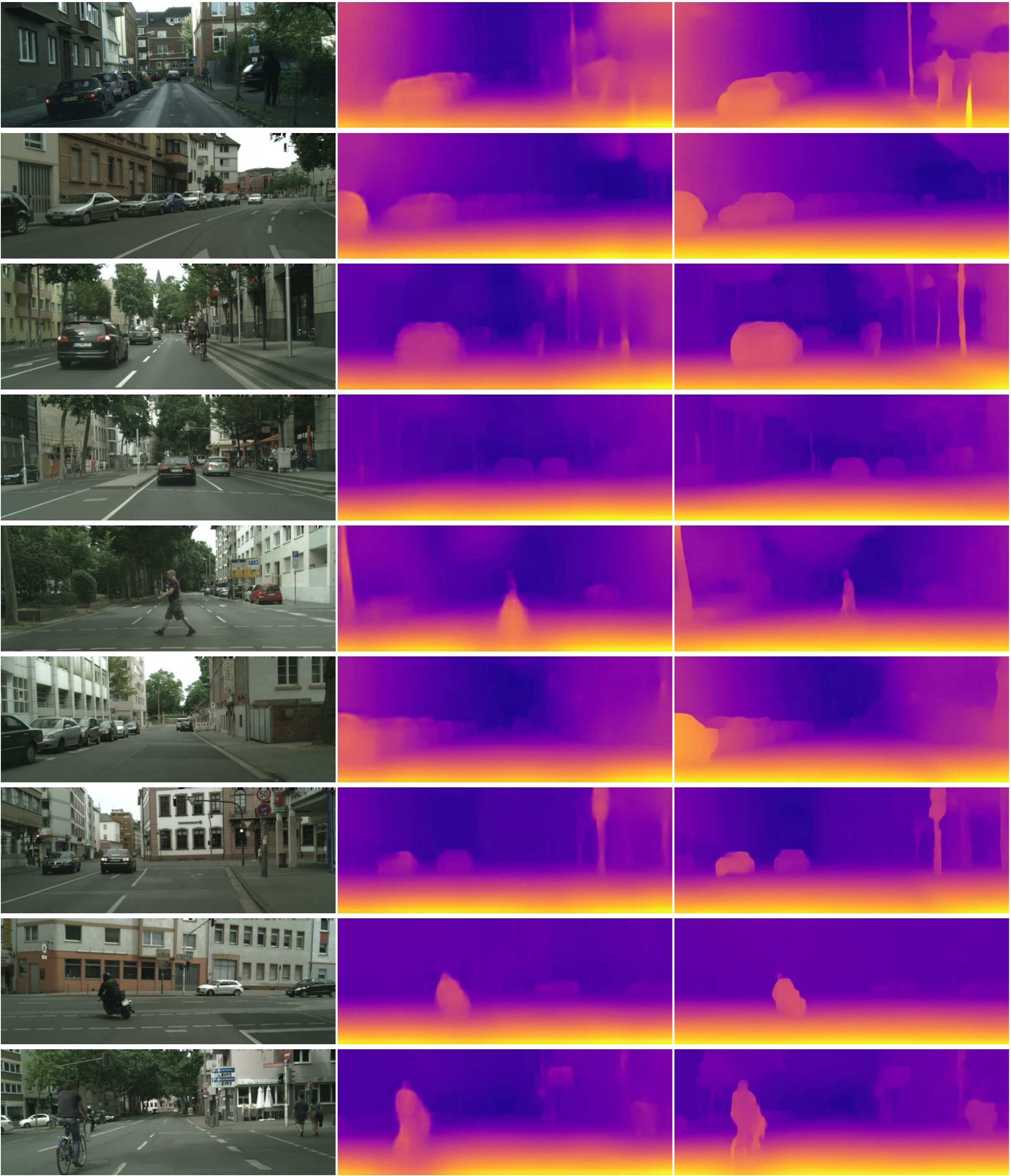}
    % \caption{}
    \label{fig: quali_1}
    % \vspace{-0.5cm}
\end{figure*}
\begin{figure*}[ht]
    \centering
    \includegraphics[width=\textwidth]{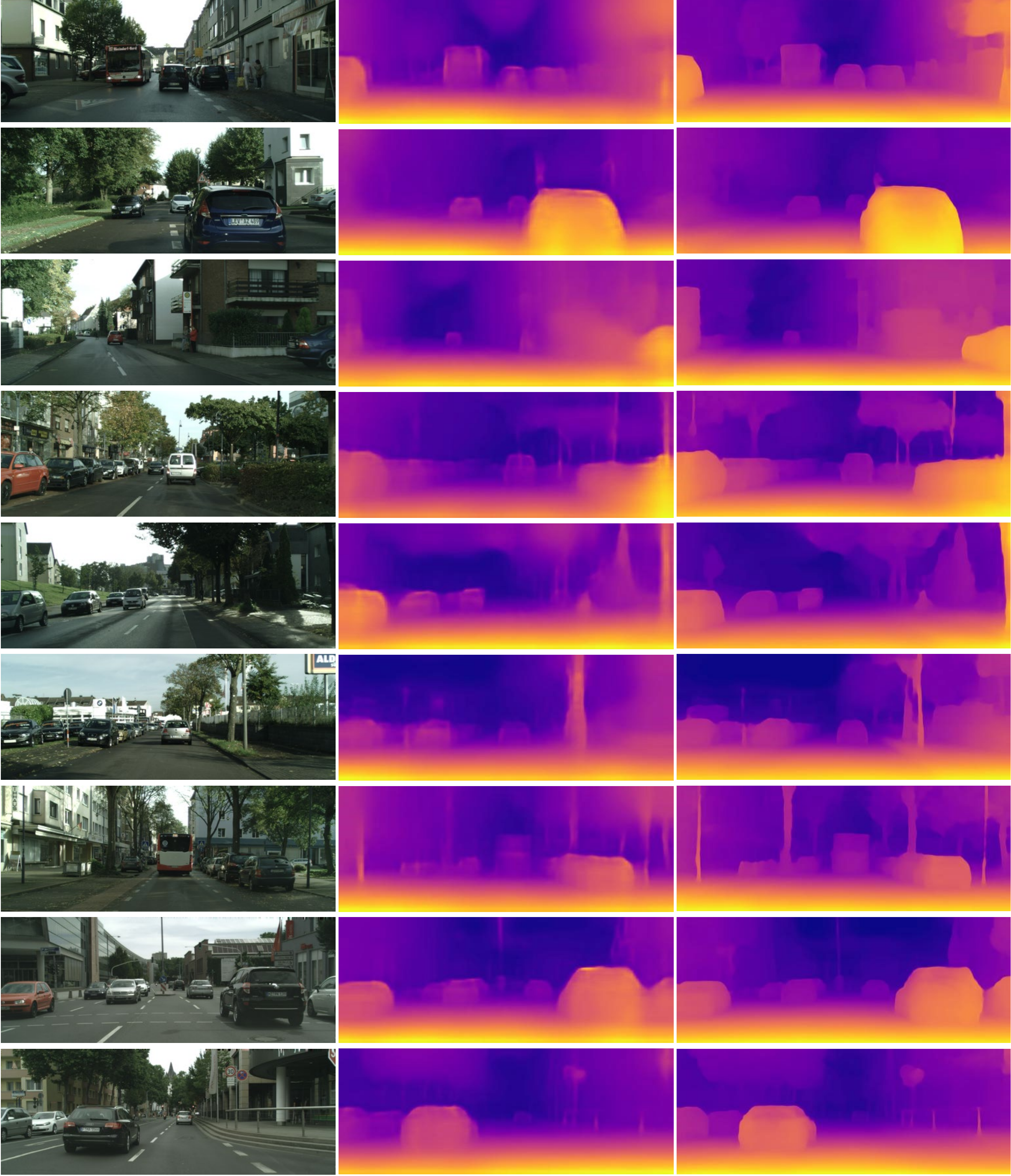}
    \caption{\textbf{Qualitative Comparisons on CityScapes test dataset}. \textit{Left}: original input; \textit{Middle}: estimated depth by full fine-tuning a U-Net from scratch; \textit{Right}: estimated depth by our PPEA-Depth.}
    \label{fig: quali_2}
    % \vspace{-0.5cm}
\end{figure*}